\documentclass[pdflatex,sn-mathphys-num]{sn-jnl}
\usepackage[english]{babel}
\usepackage[utf8]{inputenc}
\usepackage[title]{appendix}
\usepackage{comment}
\usepackage{appendix}
\usepackage{amsmath}
\usepackage{amsthm}
\usepackage{amssymb}
\usepackage{bbold}%
\usepackage{bbm,bm}
\usepackage{multirow}
\usepackage[table,xcdraw]{xcolor}
\usepackage{longtable}
\usepackage{makecell}
\usepackage{longtable}
\usepackage{caption, subcaption, float, subfloat}
\begin{document}
    
\title{Towards Optimal Valve Prescription for Transcatheter Aortic Valve Replacement (TAVR) Surgery: A Machine Learning Approach}

\author[1]{\fnm{Phevos} \sur{Paschalidis}}
\author[2]{\fnm{Vasiliki} \sur{Stoumpou}}
\author[2]{\fnm{Lisa} \sur{Everest}}
\author[2]{\fnm{Yu} \sur{Ma}}
\author[3]{\fnm{Talhat} \sur{Azemi}}
\author[3]{\fnm{Jawad} \sur{Haider}}
\author[3]{\fnm{Steven} \sur{Zweibel}}
\author[5]{\fnm{Eleftherios M.} \sur{Protopapas}}
\author[4]{\fnm{Jeff} \sur{Mather}}
\author[6]{\fnm{Maciej} \sur{Tysarowski}}
\author[5]{\fnm{George E.} \sur{Sarris}}
\author[3]{\fnm{Robert C.} \sur{Hagberg}}
\author[3]{\fnm{Howard L.} \sur{Haronian}}
\author*[2]{\fnm{Dimitris} \sur{Bertsimas}}\email{dbertsim@mit.edu}

\affil[1]{\orgdiv{J. Paulson School of Engineering and Applied Research}, \orgname{Harvard University}, \orgaddress{\city{Cambridge}, \state{MA}, \country{USA}}}
\affil[2]{\orgdiv{Operations Research Center}, \orgname{Massachusetts Institute of Technology}, \orgaddress{\city{Cambridge}, \state{MA}, \country{USA}}}
\affil[3]{\orgdiv{Heart \& Vascular Institute}, \orgname{Hartford HealthCare}, \orgaddress{\city{Hartford}, \state{CT}, \country{USA}}}
\affil[4]{\orgdiv{Hartford HealthCare Research Institute}, \orgname{Hartford HealthCare}, \orgaddress{\city{Hartford}, \state{CT}, \country{USA}}}
\affil[5]{\orgdiv{Athens Heart Surgery Institute}, \orgname{Iaso Children’s Hospital}, \orgaddress{\city{Athens}, \country{Greece}}}
\affil[6]{\orgdiv{Section of Cardiovascular Medicine}, \orgname{Yale School of Medicine}, \orgaddress{\city{New Haven}, \state{CT}, \country{USA}}}

\abstract{
    Transcatheter Aortic Valve Replacement (TAVR) has emerged as a prominent, minimally invasive treatment for patients with severe aortic stenosis, a life-threatening cardiovascular condition. Multiple transcatheter heart valves (THV) have been approved for use in TAVR, but current guidelines regarding valve type prescription remain a topic of ongoing debate within the medical community. We propose a data-driven clinical support tool to identify the optimal valve type with the objective of minimizing the risk of permanent pacemaker implantation (PPI), a predominant postoperative complication. We synthesize a novel dataset, combining U.S. and Greek patient populations, that integrates data from three distinct sources (patient demographics, computed tomography scans, echocardiograms) while harmonizing the different encoding processes specific to each country's record system. We propose leaf-level analysis to leverage the heterogeneity of the patient populations and avoid benchmarking against uncertain counterfactual risk estimates. The final prescriptive model shows a reduction in PPI rates of 26\% and 16\% compared to the current standard of care in our internal U.S. population and external, Greek validation set, respectively. To the best of our knowledge, this work represents the first unified, personalized prescription strategy for THV selection in TAVR.
}


\maketitle

\newpage
\section*{Highlights}

\begin{itemize}
    \item We develop and validate the first personalized recommendation system for transcatheter heart valve selection during Transcatheter Aortic Valve Replacement (TAVR), prescribing between Medtronic Evolut and Edwards Sapien to minimize the risk of permanent pacemaker implantation (PPI), a clinically and operationally significant complication.
    \item We train a state-of-the-art prescriptive machine learning model (Optimal Policy Trees) to learn an interpretable, unified treatment policy using a dataset synthesized from three different data sources (patient demographics, computed tomography, echocardiograms) across two countries (U.S. and Greece).    
    \item The method shows significant reduction of permanent pacemaker implantation (PPI) rate by 26\% and 16\% in the internal and external datasets relative to observed standard of care.
    \item The model’s rule-based structure enables integration into existing care pathways and aims to assist clinician decision making.
\end{itemize}

\newpage

\section{Introduction} \label{sec:int}
Transcatheter aortic valve replacement (TAVR) is a proven treatment option for patients with severe aortic stenosis across all levels of surgical risk. Since the approval of TAVR in the United States in 2011, more than 276,000 patients treatments have occurred from 2011 to 2019 and almost 73,000 in 2019 alone \cite{tavr-registry}; there is also recent expanding use of TAVR in the younger populations \cite{tavr-younger-pop}. TAVR operators replace the damaged aortic valve with an artificial transcatheter heart valve (THV); for patients not at extreme surgical risk, surgeons are limited to two platform options: the balloon-expandable Edwards Sapien platform and the self-expanding Medtronic Evolut platform \cite{zaid2023s,vinayak2024transcatheter,sethi2025novel}. Selection of valve choice by medical professionals is generally based on several factors including operator preference, patient characteristics, and valvular/annular anatomy on a computed tomography (CT) scan \cite{Mitsisi2022, leone2023prosthesis}. Crucially, the literature is still developing a clear understanding of which valve is more appropriate under which settings. Existing studies identify individual scenarios for which a specific valve may be preferred, but do not present a unified treatment strategy \cite{Mitsisi2022,leone2023prosthesis,renker2020choice}.

\hfill 

\noindent
Despite improvement in TAVR devices, implant techniques, and operator experience, permanent pacemaker implantation (PPI) {remains a frequent and clinically significant complication, associated with higher risk of long-term mortality and heart failure hospitalization \cite{wang2022predictors,siontis2014predictors,nuche2024conduction}. Operationally, post-TAVR PPI increases length of stay in the hospital and ICU and is associated with statistically significantly higher total and direct costs \cite{ahmad-ppieconomics1,brown-ppiecoinomics2}. In particular, \cite{ahmad-ppieconomics1} and \cite{brown-ppiecoinomics2} estimate that the need for PPI after TAVR increases total costs (including both inpatient and outpatient downstream care) by \$23,588 and \$10,213, respectively.

% \rev{Furthermore, PPI certainly increases the financial cost and invasiveness of care for patients receiving TAVR and should be avoided. In the medical literature, there has been previous work on the patient impact and financial costs of PPI after TAVR. For example, \cite{ahmad-ppieconomics1, brown-ppiecoinomics2} both find that patients who require PPI after TAVR had longer ICU and hospital stays, significantly higher direct and total costs, and greater resource use compared to those without PPI. Costs were consistently elevated across all categories, and PPI patients are less likely to be discharged in a timely manner. In particular, \cite{brown-ppiecoinomics2} finds statistically significantly higher costs in patients receiving PPI versus those without, including differences in total cost of \$23,588, direct costs of \$14,466, and and indirect costs of \$9,157. Similarly, \cite{ahmad-ppieconomics1} found that the need for PPI after TAVR increases total and direct costs on average by \$10,213 and \$7,087, respectively, compared to a typical TAVR admission without PPI. Given its prevalance, prior medical work and importance, as well as economic impact, we therefore focus on PPI rate as our outcome of concern in this work.}

\hfill

\noindent
Machine learning models have recently been employed to address various important diagnostic and operational problems in healthcare management.
In this work, we propose and offer an interpretable, ML model using a state-of-the-art prescriptive methodology, Optimal Policy Trees (OPT) \cite{amram2022optimal}, to prescribe THV platform choice (Medtronic Evolut or Edwards Sapien) that minimizes patients' postoperative risk of PPI. Our model is a novel, data-driven, personalized medicine approach that integrates diverse demographic, clinical, and echocardiographic data, and demonstrates an improvement of PPI rate among an internal U.S. validation dataset, as well as an external Greek validation dataset. Indeed, our proposal is the first unified treatment strategy for TAVR valve prescription, extending beyond existing studies recommending preferred valves for individuals that meet various, potentially conflicting, criteria \cite{Mitsisi2022,leone2023prosthesis,renker2020choice}.

\subsection{Related Literature}\label{subsec:related-lit}
This section discusses related work on machine learning modeling in healthcare management, causal inference methods in healthcare, and medical context on personalized TAVR treatment and PPI outcome estimations.

\hfill

\noindent
\textbf{Machine learning models in Healthcare Management} 
\hfill

\noindent
Machine learning models have become increasingly adopted into clinical practice. The improved quality of their forecasts has shown to significantly affect important operational and managerial aspects of downstream healthcare systems. Among the important outcomes include length of stay, short-term discharges, long-stay patient identification, discharge locations, flows in and out of intensive care units \cite{interpretable-inpatient-flow}, wait time \cite{Ang2016}, and patient scheduling \cite{he2019data}. Beyond traditional operational outcomes, clinical outcome driven management problems also utilize these methods for decision making, for example, developing strategies of preventative treatment allocations in diabetes \cite{kraus-prescriptive-diabetes}. Specific to cardiovascular conditions, previous works \cite{bertsimas2020personalized} have also demonstrated improved time until adverse event estimation using regression models. Our work is inline with this literature trend, and specifically focuses on the problem of cardiovasuclar conditions of TAVR.

\hfill

\noindent
\textbf{Causal Inference in Healthcare} \cite{liu-causal-inf-patient-reschedule, tuberculosis-causal-inference, wang-causal-inf-cardiovasc}.
A wide range of healthcare operations management literature has studied the problem of identifying the most effective treatment for an outcome. Specifically, Liu et al. \cite{liu-causal-inf-patient-reschedule} study the effects of rescheduling on no-show behavior in an outpatient appointment system, finding that follow-up patients are much more sensitive to rescheduling than new patients. Siddique et al. \cite{tuberculosis-causal-inference} consider the problem of multiple concurrent medication treatments and apply three causal estimation approaches (inverse probability of treatment weighting, propensity score adjustment, and targeted maximum likelihood estimation) to understand the success of using antimicrobial agents for patients with multidrug-resistant pulmonary tuberculosis. Similarly, Wang, Li, and Hopp \cite{wang-causal-inf-cardiovasc} leverage instrumental variable trees, a subset of causal trees, to evaluate outcomes across hospitals. In our work, we adopted Optimal Policy Trees \cite{amram2022optimal} which similarly computes counterfatual estimations and then globally solves for the optimal treatment for an outcome. In comparison to previous approaches on causal inference, this method provides more transparent decision paths and thus is more readily available as an interpretable support tool to the physicians.

\hfill 

\noindent
\textbf{Personalized TAVR Treatment and PPI Outcome Estimation} \cite{Mitsisi2022,leone2023prosthesis,renker2020choice,popma2019transcatheter,otto2021acc,rmilah-predict-ppi}.
Among the few existing guidelines for TAVR, most are not presented as a step-by-step decision roadmap or concrete recommendations stratified by patient characteristics. Instead, they investigate a subset of possible scenarios and apply current medical understandings of the device design and possible post-operative patient engagements to infer the choice of preferred valves \cite{Mitsisi2022,leone2023prosthesis,renker2020choice}. These approaches rely largely on qualitative assessment and are thus not evidence-based or quantitatively evaluated. Importantly, the design of their recommendations does not aim to minimize the adverse effects of a specific complication outcome. In contrast, in our case, PPI is our primary source of complication we aim to minimize. Previous medical literature has identified significant anatomical, electrocardiographic, and procedural factors such as right bundle branch block (RBBB) and bifascicular block, chronic kidney disease, male gender, diabetes mellitus, left anterior fascicular block (LAFB), as important predictors of PPI. We will leverage these existing medical knowledge to help guide the design of the machine learning models \cite{popma2019transcatheter, otto2021acc, rmilah-predict-ppi}.

\subsection{Contributions}\label{subsec:contributions}
Our contributions are as follows:
\begin{itemize}
    \item We propose the first data-driven recommendation policy for transcatheter heart valve (THV) selection in TAVR to minimize the risk of a leading post-operative complication, permanent pacemaker implantation (PPI). In comparison to existing recommendations which are often conflictive  and do not seek to optimize for a specific target (e.g., minimizing PPI), our approach provides unified, patient-specific decision rules to optimize a clinically and operationally significant objective. Leveraging our data-driven model, we also identify explicitly which factors affect the performance of different valve types.
    \item As a part of this study, we construct a novel dataset integrating two different cohorts from U.S. and Greece. This dataset, which integrates information from three distinct data sources and modalities including patient demographics, computed tomography scans and echocardiograms, harmonizes distinctive taxonomies, units, and coding processes used to describe patient clinical features across two different healthcare data record systems, and allows for an independent external validation study.
    \item We propose leaf-level analysis that is agnostic of the counterfactual predictions commonly used for model evaluation in the literature, thus comparing our policy against the observed standard of care while avoiding benchmarking against uncertain risk estimates. Leaf-level comparison also exploits the heterogeneity captured by the algorithm by allowing for analysis conducted on learned patient sub-populations. 
    \item Our model achieves significant reductions of PPI in comparison to the existing standard of care across both the internal U.S. and external Greek validation datasets: 26\% and 16\%, respectively. 
\end{itemize}

\noindent
This work is laid out in the following sections. In Section \ref{sec:methods}, we outline the methodology. This includes patient selection, an overview of the machine learning techniques we employed, such as the counterfactual estimation process and the OPT model itself, and the outline of the training of these models. Section \ref{sec:res} presents the qualitative and quantitative results of the final model, and in particular, we demonstrate the performance on the internal validation and external validation datasets. Finally, in Section \ref{sec:dis}, we provide an intensive discussion on the operational implications of our model, the clinical insights it offers, and its limitations.

\section{Method} \label{sec:methods}

In this section, we outline our methodology, beginning with a description of our patient selection and followed by an overview of the machine learning techniques involved in learning the OPTs and in quantitatively evaluating their performance.

\subsection{Patient Cohort and Characteristics} \label{sec:patientcohort}
To create our dataset, we retrospectively identify patients who received a first-time TAVR procedure between January 2011 and September 2022 at the largest US hospital network in Connecticut (Hospital A below).We merge patient demographic information with data from echocardiograms and computed tomography scans by developing code to extract features from physician reports on imaging.

\bigskip
\noindent Since valves from the Edwards Sapien and Medtronic Evolut platforms are the only valves approved for standard clinical use, we restrict attention to patients who received a valve from these platforms. Thus, all patients who did not receive a Medtronic Evolut Pro, Medtronic Evolut Pro Plus, Edwards Sapien, Edwards Sapien 3, or Edwards Sapien 3 Ultra valve were excluded from the study. Patients treated with an Edwards Sapien valve are grouped together, collectively described as having received valves from the Sapien platform. Similarly, patients who received either the Medtronic Evolut Pro or Medtronic Evolut Pro Plus valve are jointly described as having received a valve of the Evolut platform. In comparison to limiting our attention to patients who received a Sapien 3 Ultra or Medtronic Pro Plus (the most recent released versions of the respective valve platforms), these groupings allow for a larger dataset with which to train and test our model while still respecting the key differences between the Sapien and Evolut valve platforms; the former comprises balloon-expanding valves manufactured by Edwards Healthcare whereas the latter comprises self-expanding valves manufactured by Medtronic.

\hfill

\noindent
To identify patients who received pacemakers, we consider those with this complication reported in-hospital, prior to discharge, as defined in the Transcatheter Valve Therapy (TVT) Registry \cite{TVT_dictionary}. We also exclude patients who had received a pre-operative pacemaker or implantable cardioverter-defibrillator and those whose TAVR procedure was performed in a previously implanted bioprosthetic valve (“valve in valve”). Among these patients, we do not distinguish between permanent pacemaker and implantable cardioverter-defibrillators, referring to them as pacemakers collectively. This choice reflects the common properties the two devices share, since all modern implantable cardioverter defibrillators can also function as pacemakers \cite{allen2006pacemakers}. Also, as clinical outcomes, they both represent the event of emergent conduction system support through chronic pacing dependency. We additionally remove patients who had received repeated/multiple valves, and older valve types, or other miscellaneous conditions (including duplicate patient records, patients without echocardiogram data recorded, patients with more than one post-operative pacemaker implanted, or patients who did not have any complications recorded). The final study population included 1,779 patients. Figure \ref{fig:inclusion} summarizes our inclusion criteria. 

\begin{figure}[hbt]
    \centering
    \includegraphics[width=0.7\linewidth]{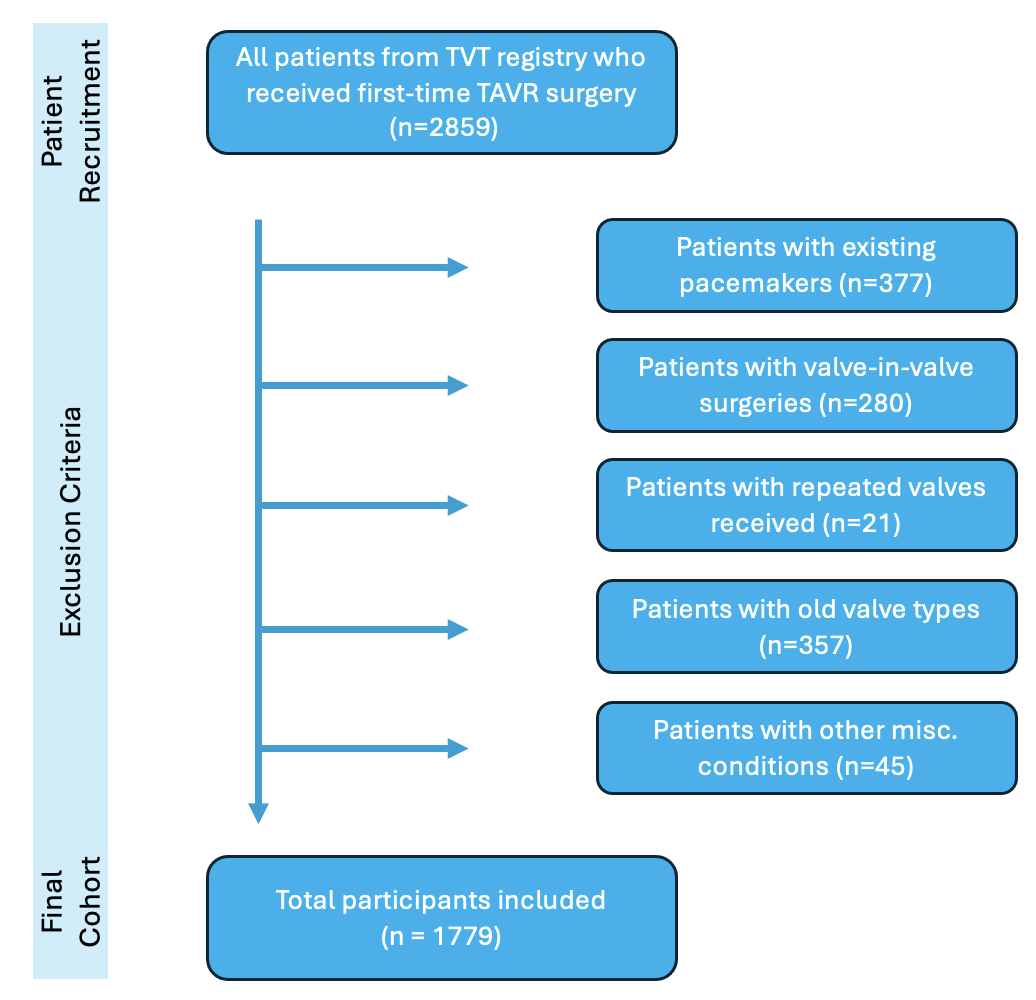}
    \caption{Inclusion criteria detailing all criterions used to define the final participating cohort}
    \label{fig:inclusion}
\end{figure}

\hfill 

\noindent
We also utilize a completely distinct patient population to form an external validation set. We retrospectively identify patients receiving a first-time TAVR procedure between January 2011 and December 2021 at Hygeia Hospital in Greece (Hospital B below) using the same criteria for exclusion as in the Hospital A population and merging data modalities similarly. We work with doctors from Hygeia Hospital and Hartford Hospital to unify the taxonomies and units of the different record systems. In this population, only two valves are represented: Edwards Sapien 3 and Medtronic Evolut Pro; no further grouping was needed. The final patient population of this external validation set is 151.

\hfill

\noindent
Table \ref{tab:valve} details the number of patients in the internal cohort that were treated with each type of valve, as well as the PPI rate among the Edwards Sapien and Medtronic Evolut patients. Table \ref{tab:valveGreek} outlines analogous statistics for the external validation set. All 47 patient features used in our study, including demographics and medical information, can be found in Appendix \ref{app:a}. More specifically, Table \ref{tab:valve_comp} in Appendix \ref{app:a} offers a more in-depth comparison between the Edwards Sapien and Medtronic Evolut patients among the internal cohort by comparing summary statistics for each of the available features. Similarly, Table \ref{tab:hospital_comp} in Appendix \ref{app:a} compares the internal and external cohorts, focusing on the most important subset of the features, as later indicated by our model.

\begin{table}[hbt]
\centering
\resizebox{0.9\textwidth}{!}{
\begin{tabular}{c|c|c|c}
Valve                     & No. Patients & Description                 & Pacemaker Rate (No.)           \\ \hline
Edwards Sapien            & 72           & \multirow{3}{*}{Edwards Sapien}   & \multirow{3}{*}{13.63\% (178)} \\
Edwards Sapien 3          & 1017         &                             &                                \\
Edwards Sapien 3 Ultra    & 217          &                             &                                \\ \hline
Medtronic Evolut Pro      & 166          & \multirow{2}{*}{Medtronic Evolut} & \multirow{2}{*}{15.22\% (72)}  \\
Medtronic Evolut Pro Plus & 307          &                             &                               
\end{tabular}}
\caption{The number of patients in the internal cohort that received each type of valve, as well as the pacemaker rate amongst the Edwards Sapien and Medtronic Evolut populations.}
\label{tab:valve}
\end{table}

\begin{table}[hbt]
\centering
\resizebox{0.9\textwidth}{!}{
\begin{tabular}{c|c|c|c}
Valve                     & No. Patients & Description   & Pacemaker Rate (No.)           \\ \hline
Edwards Sapien 3          & 118         &   Edwards Sapien     & 3.39\%  (4)                           \\
\hline
Medtronic Evolut Pro      & 33          &   Medtronic Evolut   & 18.18\% (6)   \\                          
\end{tabular}
}
\caption{The number of patients in the external validation set that received each type of valve, as well as the pacemaker rate amongst the Edwards Sapien and Medtronic Evolut populations.}
\label{tab:valveGreek}
\end{table}

\subsection{Prescriptive Methodology}\label{subsec:prescriptive}

This section focuses on the prescriptive methodology we followed for developing the final prescriptive model. 

\subsubsection{Problem Setting}
Assuming we have a dataset with $n$ observations, a formally defined prescriptive problem involves three components:

\begin{itemize}
    \item \textbf{Observational data} $\boldsymbol{x}_{i} \in \mathbb{R}^{p}$, $i=1,\ldots,n$, which refer to the known covariates (features) of our samples; each vector $\boldsymbol{x}_i$ contains the values of $p$ features and corresponds to sample $i$. These are used for the full prescriptive pipeline we employ and they include demographics, prior conditions or procedures of the patient, medications and features extracted from their echocardiograms. 
    \item \textbf{Treatments} $t \in \mathcal{T}$. We consider a set of available treatments $\mathcal{T}$, from which, for each sample $i$, one will be prescribed by the model. Treatments in our problem are discrete and refer to the two possible platforms we consider, Edwards Sapien and Medtronic Evolut, so $\mathcal{T} = \{\text{Medtronic Evolut, Edwards Sapien}\}$. 
    \item \textbf{Outcomes} $y_{i} \in \{0,1\}, \forall i=1, \dots n$. These refer to the resulting effect after applying a specific treatment. In the TAVR prescriptive problem, the outcome is binary and describes whether the patient needed a pacemaker after the TAVR procedure.
    
\end{itemize}

\hfill

\noindent
The goal of our method is to prescribe a treatment (Evolut or Sapien) to each patient, so that the risk of needing a PPI is minimized. The whole training process of the proposed pipeline consists of two separate steps: the counterfactual estimation (Section \ref{subsec:ce}) and the training of an Optimal Policy Tree that makes the final prescriptions by leveraging the calculated counterfactuals (Section \ref{subsec:prescriptive_model})

\subsubsection{Counterfactual Estimation} \label{subsec:ce}
To train a prescriptive model to make fully informed suggestions on the most appropriate treatment for each patient, it is natural to use information about the outcomes of each observation under each potential treatment. In most real-world datasets, however, only one of the available treatments is applied to each observation and the hypothetical outcomes that correspond to the remaining treatments (counterfactuals) are unknown and need to be estimated.

\hfill

\noindent
More formally, we need to know for each sample $i$ the potential outcome $y(\boldsymbol{x}_i,t)$ under all available treatments $t \in \mathcal{T}$. In the historical dataset, each sample \(i\) has been assigned to only one treatment, denoted as \(T_i\), and thus only the corresponding observed outcome (factual) is available. The counterfactuals \(y(\boldsymbol{x}_i,t)~\forall t \in \mathcal{T}\) where \(t \neq T_i\), however, remain unobserved. Consequently, the dataset contains information solely on the outcomes for the administered treatment, underscoring the necessity for counterfactual estimation to infer the outcomes under other possible treatment scenarios.

\hfill

\noindent
The approach we use for counterfactual estimation is based on causal inference literature \cite{doubly-robust-policy}. The result of this process is a counterfactual matrix $\Gamma$ of dimensions $n \times n_t$, where $n_t$ is the number of treatments (in this case $n_t = 2$). The entry $\Gamma_{it}$ corresponds to the estimated outcome for sample $i$ under treatment $t$. This matrix will be used for training our prescriptive model, as described in Section \ref{subsec:model_training}. 

\hfill

\noindent
More specifically, we use a doubly robust counterfactual estimation approach that combines a direct outcome estimator and a propensity score estimator. For the former, one model is trained for each treatment using the portions of the observational data that received the treatment. The goal of each model is to directly predict outcomes under different treatments using machine learning models, namely to learn the outcome function $y_t(\boldsymbol{x})$. A separate model is trained for each treatment $t$ by using only the subset of samples that have taken this treatment in the historical data ($\boldsymbol{x}_i, \forall i: T_i=t$). For each sample $i$, we get the estimated outcomes $\hat{y}_t(\boldsymbol{x}_i)$ for all treatments $t \in \mathcal{T}$. Models like random forests or boosted trees are typically used for this task. 

\hfill

\noindent
A weakness of the direct outcome estimation approach is the potential introduction of bias into outcome predictions, due to the possibility that data subpopulations receiving distinct treatments may be inherently different. This is often referred to as treatment assignment bias. For instance, in a binary treatment case (treatment versus no treatment), patients with more deteriorated health conditions are typically more likely to receive the treatment compared to healthier individuals. This often leads to worse observed outcomes in practice. As a result, a model trained on the subset that received the treatment may become overly pessimistic, impairing its ability to generalize effectively. In our case, as discussed in Section \ref{subsubsec:ce_qual_val}, patients in our dataset prescribed Edwards Sapien 3 and Medtronic Evolut Pro differ significantly in key features, including Area of Aortic Annulus, Left Ventricular Internal Diastolic Dimension, and Sinus of Valsalva Diameter.

\hfill

\noindent
To mitigate treatment assignment bias, one approach is to scale the estimated outcomes by the probability of each sample taking the treatment they actually received (propensity score). This probability $p_{i,t}$ of sample $i$ receiving treatment $t$ is calculated by training a classification model, called a propensity estimator, using the observational data $\boldsymbol{x}_i, \forall i: T_i=t$, that predicts treatment assignments. Again, random forests or boosted trees are employed for propensity score estimation. 

\hfill

\noindent
The doubly robust estimator uses this propensity score to re-weight the outcomes calculated through the direct outcome estimator, resulting in the final counterfactuals matrix $\Gamma$:
    
    \begin{equation}\label{eq:doubly-robust}
        \Gamma_{i,t} = \hat{y}_{i,t} + \mathbb{1}\{T_{i}=t\} \frac{1}{p_{i,t}} (y_i-\hat{y}_{i,t}).
    \end{equation}

\subsubsection{Optimal Policy Trees} \label{subsec:prescriptive_model}

Given the estimated counterfactuals, we use Optimal Policy Trees (OPT), introduced by \cite{amram2022optimal}, to learn an optimal prescription policy that minimizes postoperative pacemaker outcomes. OPT is a single-tree-based method that aims to split the input patient cohort until each subgroup is as homogeneous as possible. It is constructed by following the framework of Optimal Classification Trees \cite{oct-2017} and uses coordinate descent to optimize the objective function:

\begin{equation}\label{eq:nn_obj_func}
   \min_{\tau(.)}  \sum_{i=1}^{n} \sum_{t=1}^{n_t}\mathbb{1} \{\tau(\boldsymbol{x}_i) = t\}\cdot \Gamma_{i,t},
\end{equation}

\noindent Note that this corresponds to the sum of the estimated outcomes across all samples and can be viewed as a weighted classification tree problem, thus making the extension from the Optimal Classification Trees straightforward.

\hfill

\noindent
The tree-based methodology gives the important advantage of interpretability, since every decision the model makes to arrive at its final treatment policy is clearly documented as a binary split of patients’ characteristics. OPTs have also been shown to outperform other prescriptive methodologies \cite{amram2022optimal}. 

\subsection{Model training}\label{subsec:model_training}

The full model training pipeline consists of 3 main steps: the data preparation, the counterfactual estimation and the training of the prescriptive model. 

\subsubsection{Data Preprocessing}

As already mentioned, our tabular data $\boldsymbol{x}_i, i=1, \dots, n$ include demographics, prior conditions or procedures, medications and features extracted from echocardiograms of 1,779 patients. The full set of 47 features used can be found in tables \ref{tab:train_test_comp} and \ref{tab:valve_comp} of Appendix \ref{app:a}. 
% I switched the order of the following paragraphs

\hfill

\noindent
To study the generalizability of the model’s performance on unseen data, the data set is randomly divided into a train and test set according to a 50\%/50\% split. Table \ref{tab:train_test_comp} in Appendix \ref{app:a} offers a summary of the train and test patient cohorts. Note that this is different from the more widely accepted 80\%/20\% splits in predictive tasks since counterfactual estimation, explained in Section \ref{subsec:ce}, is separately applied to the train and test sets. Thus, to guarantee a high-quality counterfactual estimation for both sets enough data should be reserved for the test set as well.

\hfill

\noindent
Of the 1,779 patients in the internal cohort, 496 have no missing values. Thus, following the division into train and test sets, we use mean imputation to replace missing values with synthetic values meant to approximate the entries of a complete data set which has been shown to improve downstream prediction tasks \cite{bertsimas2021imputation}. More specifically, each incomplete entry of a feature is substituted with the mean of the observed values for that feature across all other samples. As shown in Figure \ref{tab:missingness}, which summarizes missingness for each feature, no single variable has more than 40\% missing data, with the vast majority mostly complete. In Appendix \ref{app:training}, we also compare our mean imputation approach with alternatives.

\hfill

\noindent
We also use an external validation set to test our results on a distinct patient population, as discussed in Section \ref{sec:patientcohort}. For this set, only patient information that is utilized by our final prescriptive model is retained. There is no missing information, and no counterfactual estimation is needed to train or evaluate the model. 

\subsubsection{Counterfactual Training}

\noindent
For the doubly robust counterfactual estimation, we train outcome and propensity estimators that are independently fit to the train and test sets. This applies to any train-test split we perform on the data. For each subset, we train two outcome estimators (one using the patients that got an Evolut valve, and one using the patients that got a Sapien valve) and one propensity estimator. After experimentation, we select XGBoost models for both estimation procedures; rewards are estimated using cross-fitting, meaning that for each observation, predictions are obtained from models trained on disjoint subsets of the data. This ensures that reward estimates are computed fully out-of-sample, mitigating overfitting and providing a more reliable assessment of model performance. The best hyperparameters for each model are chosen using cross-validation with the highest Area Under the Receiver Operating Characteristic Curve (AUC-ROC). We use a min propensity score of one when combining the outcome and propensity score estimators. In Appendix \ref{app:alt-ce}, we also explore the effect of training a single outcome estimator instead of two on the performance of the resulting prescriptive model. We observe that the results are very similar to the two-estimator pipeline; however, we select the two-estimator approach because it yields larger differences between the estimated counterfactual outcomes of the two treatments, which in turn enables clearer and more effective splitting in the trained OPT.
\subsubsection{OPT Training} \label{subsec:opt_training}
To combat overfitting, the OPTs are trained with a maximum tree depth of 8 and restricted such that each leaf contains at least 50 samples (5.6\% of training set). Moreover, 15\% of the train set is randomly withheld and used as an internal validation set in order to tune a complexity parameter that appropriately penalizes overly complicated trees. Each node in the tree indicates the number of patients in the train set that it comprises. The leaf nodes prescribe to all patients in that node a specific valve type, either Sapien or Evolut. For more information on our training approach, including hyperparameter selection, see Appendix \ref{app:training}.

\hfill 

\noindent
In order to ensure that there is no bias towards a specific train-test split and since the dataset is quite small, we train OPT models across 20 randomized train-test data splits. For each split, we first train counterfactual estimators for the train and test set separately. We then use the train set and the corresponding counterfactuals to fit multiple OPT models before evaluating their performance on the corresponding test set. In this way, we end up with trees that are trained on slightly different parts of our data.

\hfill

\noindent
The reason behind this choice lies in the fact that using a single split constrains the exploration for a meaningful and performant tree structure. After this process, the top three models are selected by balancing both performance and overfitting; based on performance in the test set that corresponds to each split, we select the top 30 models. After removing duplicates (often, training on different cohort splits results in equivalent trees) we are left with thirteen unique trees that are further pared down to three finalists based on the value and the consistency of pacemaker rate improvement in both the specific train and test sets, signaling that the model does not overfit. Such an approach is analogous to physicians with different past medical experiences making real-world decisions and can help capture the potential variability of decision-making. We then take a clinical approach to select the final OPT from these top three models; we choose the treatment strategy most consistent with the current medical insight.

\section{Results} \label{sec:res}

In the following section, we present our proposed prescriptive policy and offer in-depth analysis regarding its construction and performance. We note that throughout this section we use ``historical policy" to refer to the actual policy implemented in the data (i.e. $T_i, \forall i=1, \dots, n$), ``historical pacemaker rate" to refer to the rate of pacemaker implantation from this historical data (i.e. $\sum_{i \in \mathcal{I}} y_i / |\mathcal{I}|$ for some $\mathcal{I} \subseteq \{1, \dots, n\}$), and ``observed pacemaker rate under OPT” to refer to the rate of pacemaker implantation observed when an OPT’s prescriptions have been followed (e.g. under the prescription strategy of an OPT).

\subsection{Presentation of Proposed Policy}

\begin{figure}[hbt]
  \centering
  \includegraphics[width = 0.9\linewidth]{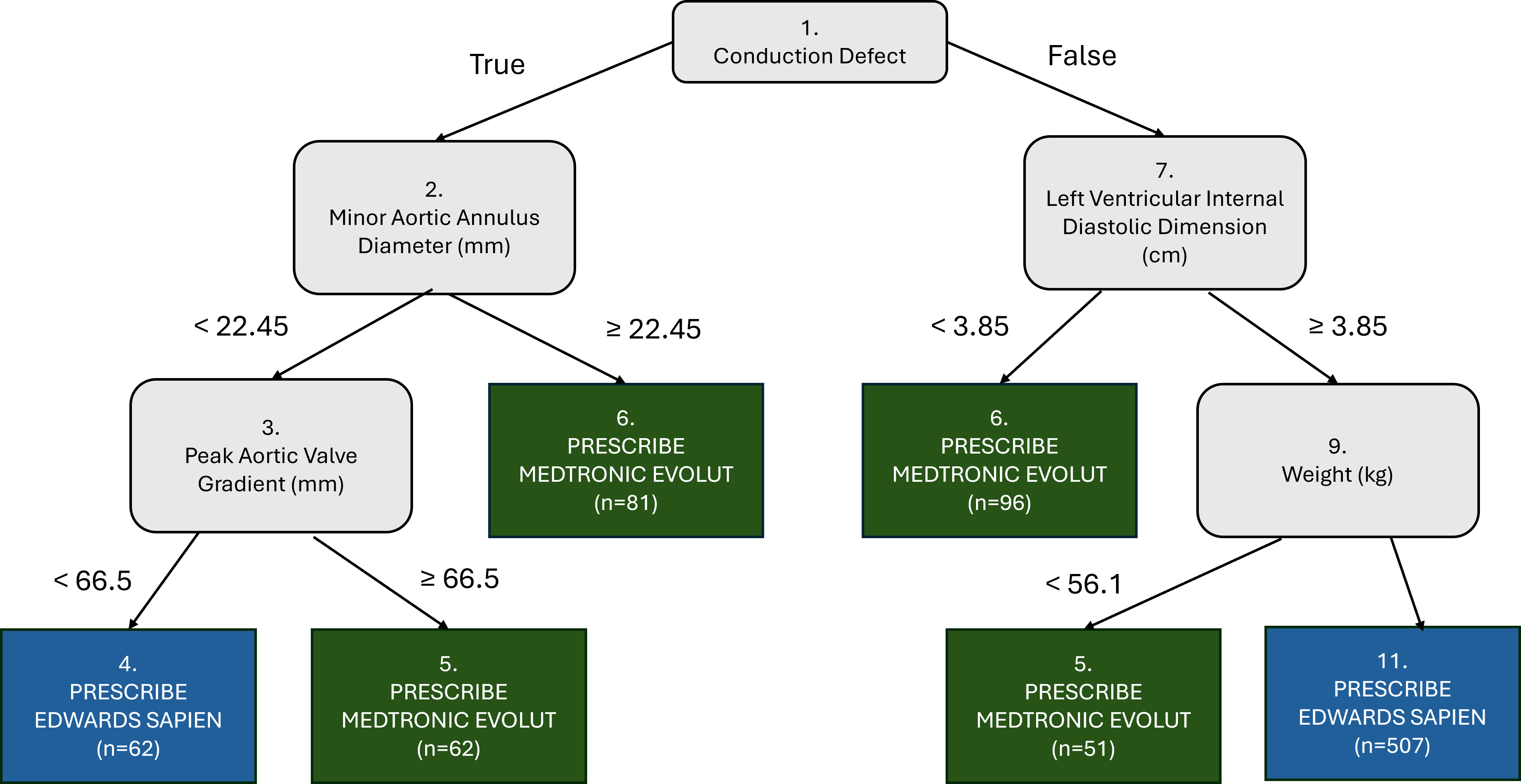}
  \caption{Proposed Prescriptive Policy: Our proposed prescriptive policy as a decision tree. The policy splits the patients on a variety of variables extracted from computed tomography scans and echocardiograms to arrive at a partition of six patient groups represented by nodes 4, 5, 6, 8, 10, and 11 and prescribes each group one of Edwards Sapien or Medtronic Evolut.}
  \label{fig:OPT}
\end{figure}

\noindent 
Figure \ref{fig:OPT} displays the OPT for optimal valve type prescription. The decision tree first splits on whether the patient has a conduction defect, defined as a left or right branch bundle block, sick sinus syndrome, or first, second, or third-degree heart block; and then on a variety of other variables: the minor aortic annulus diameter, the peak aortic valve echocardiographic gradient, the left ventricular internal diastolic dimension, and finally, patient weight. Each leaf node --- of which there are six --- states the prescribed valve and the number of patients in the train set comprising the node.

\subsection{Validation of Counterfactual Estimation} \label{sec:res:ce}

In this section, we present results related to the counterfactual estimation step of our methodology as outlined in Section \ref{subsec:ce}. Recall that our counterfactual estimation comprises six distinct models since for both the train and test set we individually train a propensity and two outcome estimators (one for Sapien platform valves and one for Evolut platform valves).

\subsubsection{Quantitative Validation}
We begin with a quantitative evaluation of the models. Table \ref{tab:reward} in Appendix \ref{app:addit_figures_counter} gives the AUC-ROC for each of these individual models. Furthermore, Figures \ref{fig:train-cc} and \ref{fig:test-cc} plot calibration curves for the Edwards Sapien and Medtronic Evolut outcome estimators for the train and test sets, respectively. In particular, for each valve type, we calibrate the train set outcome estimator for that valve platform on the testing set patients who received the valve, and vice versa. We note that the final reward estimates from the doubly robust method cannot be individually interpreted as probabilities, since the estimated outcomes are reweighted by the propensity score (see Equation \ref{eq:doubly-robust}). For this reason, we show the calibration curves for the outcome estimators before the risk predictions are reweighted.

\hfill

\noindent In Table \ref{tab:rvsp} we offer a perhaps more illustrative analysis of counterfactual performance by comparing the PPI rate estimated by the counterfactual estimators for Edwards Sapien and Medtronic Evolut patients in each leaf node to the historical PPI rates. We note that the counterfactual estimates of our models for all actually observed outcomes match the historical ground truth exactly. That is, our estimate of the pacemaker probability given Sapien of a patient who actually received Sapien and was implemented with a pacemaker is one. Beyond these actually observed outcomes, it is hard to evaluate the performance of our model on unseen counterfactuals. Nevertheless, we expect that the estimated pacemaker rates for a valve among a group of patients should be approximately equivalent to the observed rate, at least for nodes and valves with many observed patients.

\begin{table}[hbt]
\centering
\resizebox{\columnwidth}{!}{
\begin{tabular}{c|lllll}
\hline
~ & Node & \makecell[lc]{Estimated Sapien \\ Pacemaker Rate (\%)} & \makecell[lc]{Estimated Evolut \\ Pacemaker Rate (\%)} & \makecell[lc]{Historical Sapien \\ Pacemaker Rate (\%)} & \makecell[lc]{Historical Evolut \\ Pacemaker Rate (\%)} \\ \hline
\multirow{3}{16mm}{Train \newline N=890}         
& 4 & 6.07 & 17.36 & 2.5  & 22.73    \\
& 5 & 28.41 & 15.14 & 28.57 & 13.04    \\
& 6 & 35.36 & 20.44 & 37.88 & 26.67    \\ 
& 8 & 14.32 & 6.88  & 13.43 & 3.45    \\
& 10 & 15.40 & 10.88 & 15.62 & 10.53    \\
& 11 & 7.12 & 17.96 & 6.91  & 16.38    \\ 
\hline
\multirow{3}{16mm}{Test \newline N=889}          
& 4 & 19.56 & 24.48 & 21.88 & 36.84    \\
& 5 & 20.67 & 17.86 & 20.63 & 25.81    \\
& 6 & 32.16 & 18.59 & 34.21 & 7.14    \\ 
& 8 & 5.43 & 11.17  & 4.84  & 5.0    \\
& 10 & 4.55 & 8.53  & 3.03  & 0.00    \\
& 11 & 8.98 & 15.48 & 9.94  & 14.58    \\ 
\end{tabular}
}
\caption{Performance of Counterfactual Estimation: The estimated pacemaker rate of Edwards Sapien and Medtronic Evolut at each node among the train and test set as conjectured by the corresponding counterfactual estimator compared to the historical pacemaker rate amongst Edwards Sapien and Medtronic Evolut patients at the node. }
\label{tab:rvsp}
\end{table}

\hfill

\noindent We note that the estimated pacemaker rate of the train and test sets at each node mostly approximates the historical pacemaker rate among the corresponding population. Indeed, the counterfactual estimation model built on the training set shows remarkably consistent approximation, with the biggest deviation coming from the estimated and historical pacemaker rates of the Evolut patients in Node 6. The testing set model exhibits larger discrepancies, but, critically, its estimates preserve the more successful valve (the valve which observed the lower historical pacemaker rate) for a majority of nodes, including nodes 4 and 6. The only exceptions are in the counterfactual estimation of the Evolut pacemaker rate for test set patients in nodes 5, 8, and 10, which are perhaps explained by the small size of these patient populations (there were just 31, 20, and 21 test set patients in nodes 5, 8 and 10, respectively, who received an Evolut). Importantly, the test set counterfactuals are not used in model training, but rather as one method for evaluating the final OPT (see Section \ref{sec:main-analysis}). 

\subsubsection{Qualitative Validation} \label{subsubsec:ce_qual_val}

We also validate our counterfactual estimators qualitatively through analysis of their feature importances. Figure \ref{fig:estimate-feat-import} visualizes the ten most important features to the Edwards Sapien outcome estimators by averaging over the train and test set models. An analogous plot for the Medtronic Evolut estimator is deferred to Appendix \ref{app:addit_figures_counter}. Given that the goal of the outcome estimators is to predict risk of pacemaker for the Sapien and Evolut platforms, we expect that many of these features have been identified in the medical literature as risk factors for PPI following TAVR. Indeed, conduction defect (especially left and right branch bundle blocks), age, and weight have all been associated with need for PPI for both balloon-expandable and self-expanding valves \cite{rudolph2023modifiable}. Furthermore, the anatomic variables featured such as Peak Aortic Valve Gradient, Aortic Valve Area, and Minor Aortic Annulus Diameter may be proxies for information about the extent of aortic valve calcification, another key risk factor for PPI \cite{rudolph2023modifiable, gonska2017predictors}.

\begin{figure}[hbt]
    \centering
        \includegraphics[width=\linewidth]{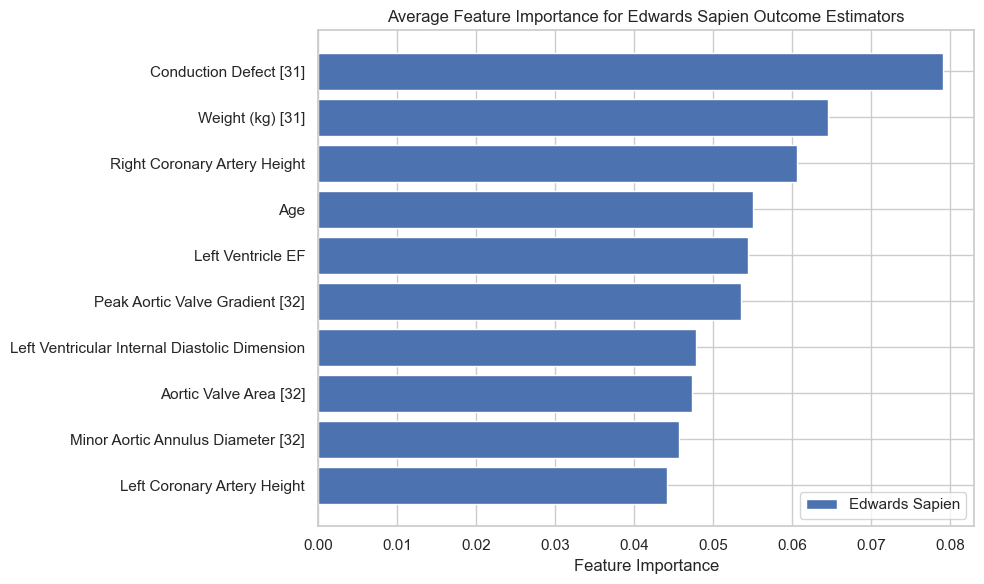}
        \label{fig:estimate-feat-import-sapien}    
    \caption{Sapien Feature Importance: Feature importances of the Edwards Sapien valve outcome estimators. Importance scores are averaged over the associated train and test set models. Note that the absolute importance of any one variable has no interpretation. Instead, it is the relative difference in feature importances that offer insight into which variables are used by the model.}
    \label{fig:estimate-feat-import}
\end{figure}

\hfill

\noindent For the propensity estimators, we similarly average feature importances for the train and test set models to identify the most important variables. Figure \ref{fig:ce-propensity} compares the standardized and truncated distributions of Sapien and Evolut patients for these features. We observe that many of these features such as Area of Aortic Annulus, Left Ventricular Internal Systolic Dimension, Sinus of Valsalva Diameter, and Major Aortic Annulus Diameter, have distributions that are noticeably different for the two patient cohorts, indicating that they can indeed be used to differentiate between cohorts historically prescribed Sapien and Evolut platforms valves. Indeed, by performing a simple independent $t$-test (significance threshold $p_{\text{sig}} = 0.05$) comparing the distribution of each of the ten most important features among Edwards Sapien and Medtronic Evolut patients, we find that six of the ten features have significantly different distributions among the two cohorts (Area of Aortic Annulus, $p < 0.001$; Left Ventricular Internal Diastolic Dimension, $p=0.002$; Left Ventricular Internal Systolic Dimension, $p<0.001$; Aortic Root Angle, $p=0.001$; Sinus of Valsalva Diameter, $p<0.001$, Major Aortic Annulus Diameter, $p<0.001$). Though there exists limited literature retrospectively identifying differences in patient populations fitted with Sapien and Evolut valves, personalized recommendation approaches \cite{Mitsisi2022,leone2023prosthesis,renker2020choice} advise using balloon-expanding Sapien valves on patients with larger aortic annulus sizes and self-expanding Evolut valves on patients with smaller ones, potentially explaining the statistically significant differences observed in key variables like Area of Aortic Annulus and Major Aortic Annulus Diameter seen in Figure \ref{fig:ce-propensity}. We briefly note that these significant differences further justify our use of doubly robust estimation, as discussed in Section \ref{subsec:ce}.

\begin{figure}[!]
    \centering
    \includegraphics[width=\linewidth]{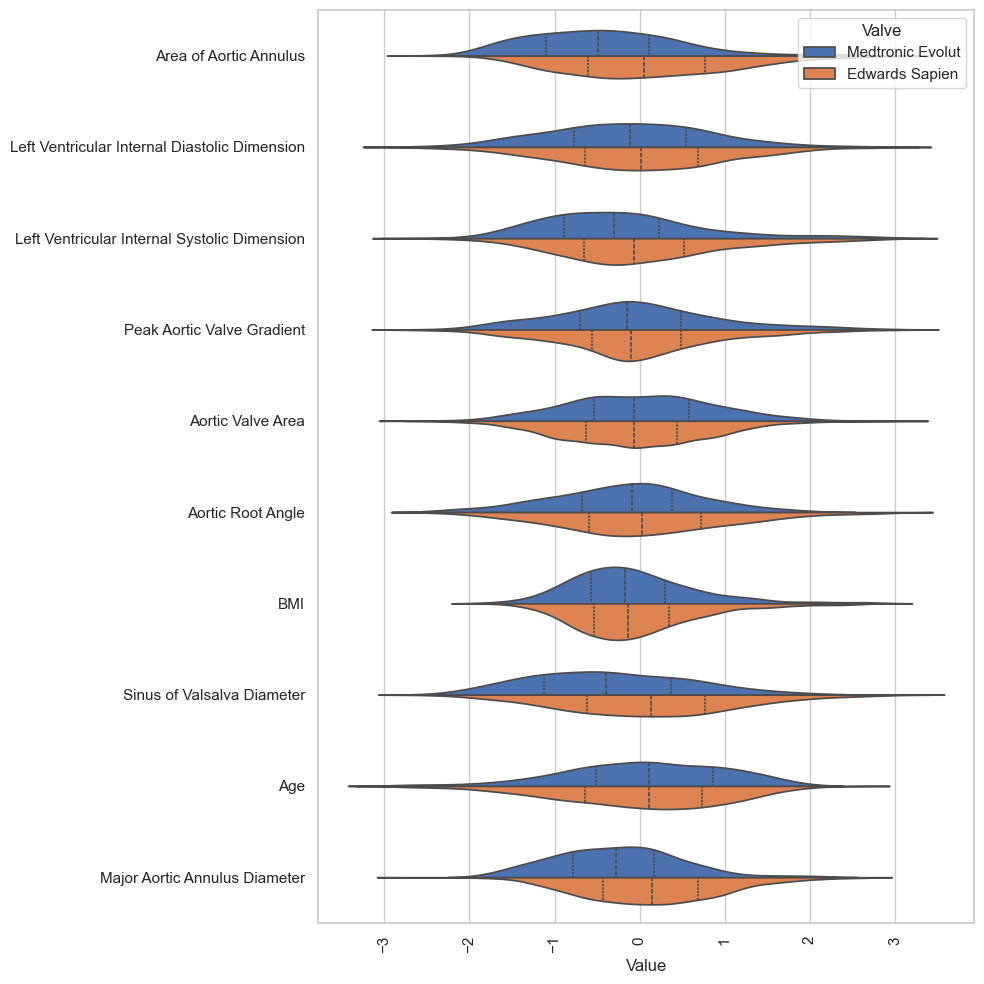}
    \caption{The standardized distributions of patients prescribed the Edwards Sapien and Medtronic Evolut platforms with tails cut-off after three standard deviations for each of the ten most important features for the propensity estimators. Feature importance scores are averaged over the train and test set models.}
    \label{fig:ce-propensity}
\end{figure}

\subsection{Analysis of Proposed Policy} \label{sec:main-analysis}

\noindent Turning to the OPT itself, Table \ref{tab:nodes} details the historical pacemaker rate at each of the six leaf nodes among patients fitted with the Sapien and Evolut platforms valves in each of the three sets (total, train, and test) as well as the valve prescribed by the OPT. Note that only for nodes 5 and 8 does the OPT mistakenly prescribe the valve that performs worse in the test set; the right valve is always prescribed for the train set. Given that the more ``successful'' valve for these nodes (the valve with the smaller pacemaker rate) is different in the training and testing sets, this discrepancy makes intuitive sense; the model selected the valve that performed better on patients it had seen. Ideally, the splits would be correctly optimized such that the successful valve remains consistent from training to testing among each node, but we note that both nodes 5 and 8 comprise a small percentage of our total population and have similar pacemaker rates for Sapien and Evolut patients over the total population (that is, combining the testing and training sets). As such, as we will soon see, this discrepancy does not seriously affect the performance of our model on the test set. 

\hfill

\noindent
To quantify the improvement in pacemaker rate associated with this appealing prescription pattern we assume that, at any node, if the patients who received an Evolut valve were instead fitted with Sapien, they would observe the same pacemaker rate as the patients who did receive a Sapien valve and vice versa. We then calculate the observed pacemaker rate under our prescribed policy for each set. The results are illustrated in Table \ref{tab:overv}, which shows that the OPT offers an improvement in both the train and the test set, decreasing the pacemaker rate by close to 4\% in the combined patient population which amounts to a 26.5\% reduction in the need for PPI. We also observe that the differences between train and test set performances are relatively close in percent improvement, meaning that our model did not overfit or underfit and can generalize well in unseen datasets. In Section \ref{subsubsec:external-validation}, we further investigate this generalizability by evaluating the performance of the OPT on an external validation dataset. Additionally, in Appendix \ref{app:addit_figures_subgroup}, we perform a subgroup analysis of our model on patients of different age ranges to isolate populations for which our model is more or less performant.

\hfill 

\begin{table}[hbt]
\centering
\resizebox{\textwidth}{!}{
\begin{tabular}{c|llllll}
\multicolumn{1}{l|}{} &
    Node &
    \makecell[lc]{Prescribed Treatment} &
    \makecell[lc]{No. Implanted \\ Edwards Sapiens} &
    \makecell[lc]{Historical Sapien \\ Pacemaker Rate (\%)} &
    \makecell[lc]{No. Implanted \\ Medtronic Evoluts} &
    \makecell[lc]{Historical Evolut \\ Pacemaker Rate (\%)} \\ \hline
\multirow{6}{16mm}{Total \newline N=1779} & 4  & Edwards Sapien   & 104 & 14.42 & 41  & 29.27 \\
                                         & 5  & Medtronic Evolut & 133 & 24.81 & 54  & 20.37 \\
                                         & 6  & Medtronic Evolut & 142 & 35.92 & 29  & 17.24 \\
                                         & 8  & Medtronic Evolut & 129 & 9.3   & 49  & 4.08  \\
                                         & 10 & Medtronic Evolut & 65  & 9.23  & 40  & 5.0   \\
                                         & 11 & Edwards Sapien   & 733 & 8.32  & 260 & 15.38 \\ \hline
\multirow{6}{16mm}{Train \newline N=890} & 4  & Edwards Sapien   & 40  & 2.5   & 22  & 22.73 \\
                                        & 5  & Medtronic Evolut & 70  & 28.57 & 23  & 13.04 \\
                                        & 6  & Medtronic Evolut & 66  & 37.88 & 15  & 26.67 \\
                                        & 8  & Medtronic Evolut & 67  & 13.43 & 29  & 3.45  \\
                                        & 10 & Medtronic Evolut & 32  & 15.62 & 19  & 10.53 \\
                                        & 11 & Edwards Sapien   & 391 & 6.91  & 116 & 16.38 \\ \hline
\multirow{6}{16mm}{Test \newline N=889}   & 4  & Edwards Sapien   & 64  & 21.88 & 19  & 36.84 \\
                                        & 5  & Medtronic Evolut & 63  & 20.63 & 31  & 25.81 \\
                                        & 6  & Medtronic Evolut & 76  & 34.21 & 14  & 7.14  \\
                                        & 8  & Medtronic Evolut & 62  & 4.84  & 20  & 5.0   \\
                                        & 10 & Medtronic Evolut & 33  & 3.03  & 21  & 0.0   \\
                                        & 11 & Edwards Sapien   & 342 & 9.94  & 144 & 14.58 \\ \hline
\end{tabular}
}
\caption{Historical Outcome Performance: The size and PPI rate of the historical Edwards Sapien and Medtronic Evolut patient implants in (i) the combined train and test set, (ii) the train set, and (iii) the test set for each node, as well as the treatment prescribed by the OPT.}
\label{tab:nodes}
\end{table}

\begin{table}[hbt]
\centering
\resizebox{\textwidth}{!}{
\begin{tabular}{cccc}
\hline
       &  Historical Pacemaker Rate  &  Observed Pacemaker Rate under OPT  &  Percent Improvement  \\
\hline
 Total &         14.05\%          &               10.32\%                &        26.54       \\
 Train &          13.6\%          &                8.87\%                &        34.74\      \\
 Test  &         14.51\%          &               11.39\%                &         21.5       \\
\hline
\end{tabular}
}
\caption{Pacemaker Rate Improvement: The historical pacemaker rate of the total patient population, the train set, and the test set compared to the pacemaker rate observed under the OPT's prescription strategy.}
\label{tab:overv}
\end{table}

\begin{table}[hbt]
\centering
\resizebox{\textwidth}{!}{
\begin{tabular}{cccc}
\hline
       &  Actual Pacemaker Rate  &  Observed Pacemaker Rate under OPT  &  Percent Improvement  \\
\hline
 Train &          13.6\%          &                9.29\%                &         31.7      \\
 Test  &         14.51\%          &               12.05\%                &        16.92       \\
\hline
\end{tabular}
}
\caption{Pacemaker Rate Improvement (CE): The historical pacemaker rate of the training and tests set compared to the pacemaker rate of the policy prescribed by the OPT as evaluated by the counterfactual estimators.}
\label{tab:rewardestim}
\end{table}

\noindent
We can also evaluate the model using the counterfactual estimators discussed in Sections \ref{subsec:ce} and \ref{sec:res:ce}. The results, summarized in Table \ref{tab:rewardestim}, offer more evidence that the policy prescribed by the OPT significantly outperforms the existing treatment strategy. It should also be noted that the actual relative improvement percentages (31.7\% and 16.92\% for train and test set respectively) are similar to the percentages obtained by the node-based analysis of Table \ref{tab:overv}.

\hfill

\noindent Finally, through bootstrap samples of the test set, we can create confidence intervals around the estimated improvement reported in Tables \ref{tab:overv} and \ref{tab:rewardestim}. In particular, we create $B=1000$ bootstrap samples of 95\% of the test set (sampled with replacement) and evaluate the observed improvement in pacemaker rate through the node analysis technique of Table \ref{tab:overv} and through the trained counterfactual estimators. Figure \ref{fig:ci} shows the resulting distribution over the 1000 trials; the 95\% confidence interval of the node analysis and counterfactual evaluation techniques are (0.037, 0.378) and (0.0395, 0.279) respectively. Outliers associated with the node analysis technique are probably a result of bootstrapped samples where some nodes comprise few patients, thus skewing the historical pacemaker rate we use to calculate the observed pacemaker rate under the OPT. Even so, both evaluation techniques show that our proposed policy outperforms the historical policy with high confidence.

\begin{figure}[hbt]
    \centering
    \includegraphics[width=0.8\linewidth]{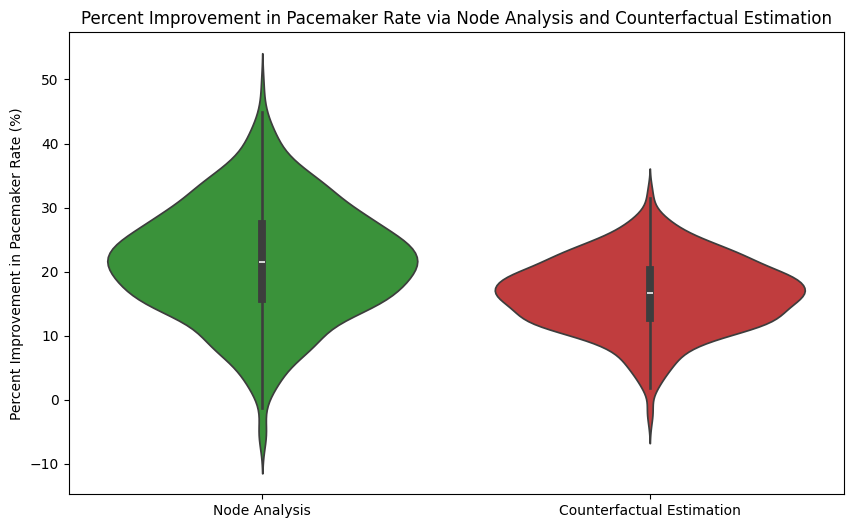}
    \caption{The percent improvement in pacemaker rate induced by our proposed policy over 1000 bootstrap test sets as evaluated by our node analysis technique and by the counterfactual estimators. The means are denoted by white lines. The 95\% confidence intervals of the techniques are (0.036, 0.377) and (0.0646, 0.275) respectively.}
    \label{fig:ci}
\end{figure}

\subsection{Performance on External Dataset}\label{subsubsec:external-validation}
We perform a similar analysis on an external cohort. Table \ref{tab:ood} lists the real pacemaker rate of each leaf node as well as the valve prescribed by the OPT. In analyzing the performance of the OPT, it is important to note that in many of the leaf nodes there are too few patients to draw conclusions about the model’s prescription policy. However, in Node 11, where there are significant amounts of both Edwards Sapien and Medtronic Evolut patients, the OPT prescribes the valve with a significantly lower rate of PPI. Since Node 11 is also the most populous node in the internal cohort, it is particularly important that the OPT generalizes well for the patients assigned to that node. Table \ref{tab:overvood} compares the historical pacemaker rate of the out-of-sample population with the pacemaker rate observed under the OPT. Despite this set observing a significantly smaller true pacemaker rate in comparison with the internal cohort, the OPTs prescription strategy still offers a significant improvement. The 16.62\% percent reduction is comparable to the 21.5\% reduction observed in the test set.

\begin{table}[hbt]
\centering
\resizebox{\textwidth}{!}{
\begin{tabular}{c|llllll}
\hline
~ &
    Node &
    \makecell[lc]{Prescribed Treatment} &
    \makecell[lc]{No. Implanted \\ Edwards Sapiens} &
    \makecell[lc]{Historical Sapien \\ Pacemaker Rate (\%)} &
    \makecell[lc]{No. Implanted \\ Medtronic Evoluts} &
    \makecell[lc]{Historical Evolut \\ Pacemaker Rate (\%)} \\ \hline
\multirow{3}{16mm}{Greek \newline N=151}   
& 4  & Edwards Sapien    & 5 & 0.00   & 0 & 0.00    \\
& 5  & Medtronic Evolut    & 5 & 0.00   & 4 & 0.00    \\
& 6  & Medtronic Evolut    & 4 & 25.00   & 3 & 3.33    \\
& 8  & Medtronic Evolut    & 12 & 0.00   & 2 & 0.00    \\
& 10 & Medtronic Evolut   & 8 & 0.00  & 3 & 0.00    \\
& 11 & Edwards Sapien     & 84 & 3.57 & 21 & 19.05    \\ \hline
\end{tabular}
}
\caption{External Cohort Performance: The size and PPI rate of the actual Edwards Sapien and Medtronic Evolut patient implants in the external validation set as well as the treatment prescribed by the OPT.}
\label{tab:ood}
\end{table}

\begin{table}[hbt]
\centering
\resizebox{\textwidth}{!}{
\begin{tabular}{cccc}
\hline
      & Actual Pacemaker Rate & Observed Pacemaker Rate under OPT &  Percent Improvement\\ 
      \hline
Greek & 6.62\%                & 5.52\%     & 16.62
\\ \hline
\end{tabular}
}
\caption{External Cohort Improvement: The historical pacemaker rate of the training, and the test set as well as the pacemaker rate of the policy prescribed by the OPT as evaluated by the counterfactual estimators.}
\label{tab:overvood}
\end{table}

\hfill

\noindent
These results demonstrate the viability of the model on an unseen cohort with important differences to the patient population on which it was trained. As discussed in Section \ref{subsec:opt_training}, the results of both the training and testing sets were used to evaluate and select our proposed policy. Its performance on an external validation set, therefore, is important to affirm that our final model does not perform disproportionately well on its training and testing set as a byproduct of our model selection process. Furthermore, we note that though the external validation set's patient population is similar in pre-operative characteristics (see Table \ref{tab:hospital_comp}), the outcomes of the TAVR surgeries are quite different. As Table \ref{tab:valve} and \ref{tab:valveGreek} demonstrate, the historical pacemaker rate of surgeries with Edwards Sapien and Medtronic Evolut platforms valves are notably different than the analogous procedures at participating U.S. Hospital.

\subsection{Comparison to Other Candidate Models}

Finally, we present results related to our model selection process as outlined in Section \ref{subsec:model_training}. In particular, we seek to understand our model's relationship to the other twelve best-performing OPTs, and especially the two other finalists, in both treatment policy and actual performance.

\hfill

\noindent
We begin by calculating the concordance of our model and the other twelve best-performing OPTs, including the other two finalists (Fin1 and Fin2), where we define concordance as the fraction of patients for which both models prescribe the same treatment. As shown in Table \ref{tab:conc}, our model agrees on more than 70\% of all patients with all other high-performing OPTs, pairwise. Similarly, we further calculate that our model and the other two finalists agree unanimously on 70.4\% of all patients. Thus, among performant OPTs, and especially among the trees which we distinguished only through a clinical approach, we see consistent agreement on a majority of patients.}

\begin{table}[hbt]
\centering
\resizebox{\textwidth}{!}{
\begin{tabular}{cccccccccccc}
\hline
Fin1 & Fin2 & OPT1 & OPT2 & OPT3 & OPT4 & OPT5 & OPT6 & OPT7 & OPT8 & OPT9 & OPT10 \\ \hline
0.81 & 0.72 & 0.75 & 0.72 & 0.85 & 0.71 & 0.75 & 0.74 & 0.74 & 0.72 & 0.78 & 0.74
\\ \hline 
\end{tabular}
}
\caption{Concordance between our chosen model and the twelve other high-performing OPTs, including the other finalists, Fin1 and Fin2.}
\label{tab:conc}
\end{table}

\noindent
To evaluate also the difference in performance across our model pool, we list the percent improvement of each of our other twelve candidates on their respective test sets (see Table \ref{tab:comp-test-res}). As expected, our chosen model (recall, improvement 21.5\%) outperforms all other candidates. Nevertheless, we see that all models in our pool demonstrate some improvement over the historical policy with the other two finalists in particular reaching similar levels as our chosen model. For completeness, the full results for these two finalists can be found in Appendix \ref{app:alt-models}.

\begin{table}[hbt]
\centering
\resizebox{\textwidth}{!}{
\begin{tabular}{cccccccccccc}
\hline
Fin1 & Fin2 & OPT1 & OPT2 & OPT3 & OPT4 & OPT5 & OPT6 & OPT7 & OPT8 & OPT9 & OPT10 \\ \hline
16.36\% & 16.96\% & 6.33\% & 5.49\% & 6.01\% & 3.52\% & 16.46\% & 11.35\% & 9.14\% & 16.86\% & 10.39\% & 10.8\%
\\ \hline 
\end{tabular}
}
\caption{Estimated improvement (as compared to historical policy) of other trees on their respective test sets. Recall that the estimated improvement of our chosen model was 21.5\%.}
\label{tab:comp-test-res}
\end{table}

\hfill

\noindent
We can also quantify the estimated improvement in pacemaker rate of the other finalists on the same test set used for our final model. Since each model was trained on a different split of data, we note that some of the patients on the test set of our final model will have been seen as training examples for other models. Nevertheless, we believe comparing model performance on the same cohort of patients is perhaps more insightful than just evaluating performance of each model on its own test set, especially since the latter metric was in part used during the selection process. In particular, we find that Fin1 observed an estimated improvement of 13\%, and Fin2 an estimated improvement of 27\%.

\subsection{Interactive, Publicly Available Prescription Calculator}
\begin{figure}
    \centering
    \includegraphics[width=0.5\linewidth]{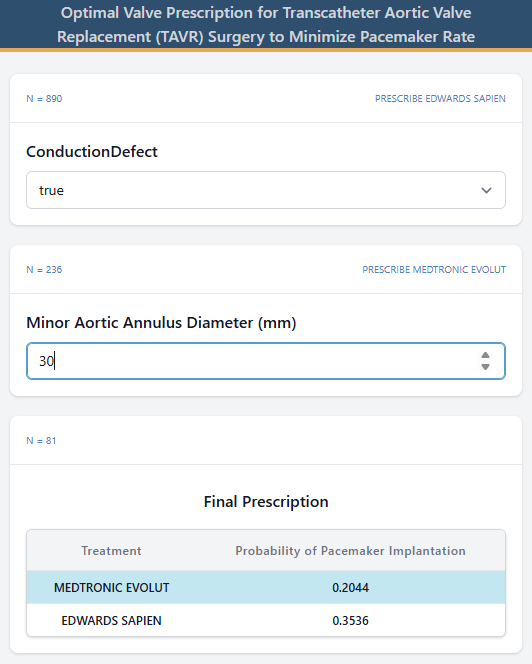}
    \caption{Visualization of the online algorithm user interface}
    \label{fig:enter-label}
\end{figure}

We demonstrate the feasibility of integrating the proposed tree model into an interactive, publicly available prescription calculator (available \href{https://raw.githack.com/yuma-sudo/TAVR-App/main/Optimal\%20Valve\%20Type\%20for\%20TAVR.html}{here}) that can be easily adopted either in physicians' mobile devices or existing hospital infrastructures, as shown in Figure \ref{fig:enter-label}. The calculator provides different risk estimations of pacemaker implantation given different valve choices and allows physicians and patients to interpret and understand the decision-making process of the model.

\section{Discussion} \label{sec:dis}
\subsection{Operational Implications for Patients and Hospital Entities}
Widely used cardiovascular risk calculators are predominantly used as tools to predict estimates of complications following the TAVR surgery. However, these tools are offered a posteriori to evaluate the patient’s condition. A significant advantage of the proposed prescriptive AI method is to offer interventions prior to TAVR surgery to directly affect outcomes. This approach shifts the focus from reactive to proactive care, allowing healthcare providers to make informed decisions before surgery rather than merely assessing risks afterward. 

\hfill 

\noindent
The decision and procurement process of valve prescription prior to the surgery is complex and cost consuming. After the patient undergoes comprehensive evaluation, often involving a set of imaging studies, a team of heart experts consisting of interventional cardiologists, cardiac surgeons, and others must convene to discuss the findings and make a final, collective decision on the most appropriate valve. However, these processes often involve a large amount of patient-specific data on anatomical, clinical, and electrocardiographic factors in a short period of time, and thus it remains challenging for physicians to truly incorporate all aspects of information simultaneously. In addition, the valve choice still can be influenced by operator bias such as their past experience with a particular valve type, thus contributing to sub-optimal valve type selection for patients. As of now, there are no clear clinical guidelines or recommendations on how to best integrate clinical judgement, patient data, and operational considerations to optimize valve prescription. Notably, our automated ML-based prescriptive method is the first such proposed guideline, with concrete decision pathways and data-driven evidence for physicians to optimally select valve choice for each individual patient.

\hfill 

\noindent
Given its more accurate valve choice recommendations, our algorithm can further improve downstream operational efficiency by allowing surgical planning team to prepare for TAVR surgery more readily; by setting up appropriate delivery surgical systems and equipment, the proposed tool can potentially reduce procedure time in the operating room and lower complications associated with prolonged surgery, such as infection or bleeding.

\hfill 

\noindent
Most importantly, evidence from both the U.S. cohort and the external Greek cohort indicates a significantly reduced rate of permanent pacemaker implantation, at 26\% and 16\% respectively. This would materially improve patient quality of life in part by shortening the length of hospital stay attributed to additional post-operative patient monitoring and recovery. From the hospital perspective, this allows quicker patient turnover to free up hospital beds, thus reducing potential ingestions in patient flow. The financial implication of the reduced risk is also significant. With average costs ranging from \$10,000 to \$50,000 pre-insurance, and \$2800 to \$4500 post-insurance, PPI imposes a serious financial barrier and unfairly penalizes demographics of underprivileged backgrounds. In addition, costs related to follow-up procedures during readmissions can also pose challenges for both patients and hospitals to manage and can introduce further unnecessary constraints on critical hospital resources.

\subsection{Clinical Insights}
The transparent, interpretable decision pathways of the tree-based models also offer some important clinical insights. Numerous pre-procedural, procedural and valve related factors have been shown to be predictive of PPI such as pre-existing conduction defects, diabetes mellitus, and chronic kidney disease \cite{popma2019transcatheter, otto2021acc, rmilah-predict-ppi}. In addition, left bundle branch block (LBBB) and high grade atrioventricular (AV) block are common post TAVR surgery \cite{lee2015mechanisms}. However, our study shows that the occurrences of these defects are not only predictors of postoperative PPI or common complications, but can be used as factors to influence choice of valve type.

\bigskip
\noindent In comparison with prior work on valve prescription \cite{Mitsisi2022,leone2023prosthesis,renker2020choice}, our model also identifies the presence of a conduction defect and the size of aortic annulus as factors that can affect the performance of the different valve types. Other features used by our model, such as Peak Aortic Valve Gradient and Left Ventricular Internal Diastolic Dimension may be proxies for other factors already identified by the literature, such as Aortic Valve Calcification and Left Ventricular Ejection Fraction (LVEF). Our unified treatment approach is different than the more qualitative, scenario-based discussion of previous works, which makes it hard to compare recommendations. For example, Mitsisi et al. and Leone et al. both recommend self-expanding valves like Sapien 3 for patients with large annulus sizes \cite{Mitsisi2022,leone2023prosthesis} In our model, when splitting on Minor Aortic Annulus in Node 2 of the tree (see Figure \ref{fig:OPT}), we prescribe Evolut for patients with Minor Aortic Annulus Diameter $\geq 22.45$mm. Nevertheless, this split is only for patients that also have a pre-existing conduction defect, and thus does not necessarily contradict the suggestions of the existing literature \cite{Mitsisi2022,leone2023prosthesis}.

\subsection{Limitations}
We note a few limitations of the study. Our model training patient population is not representative of the general U.S. population as it comes from a single hospital. Although we have indicated the model’s generalizability using an external Greek validation dataset, further external testing in other institutions should precede the application of the proposed model in different healthcare settings. We also mention that there is room for improvement in model performance, although our model is estimated to outperform the existing standard of care. Future studies should consider including additional features known to affect complication outcomes, including anatomic/morphometric information regarding the LVOT, the aortic valve, the coronary arteries, the calcium load, its extent, and distribution. In addition, given the emergence of a wealth of multimodal learning literature, we note the potential to include a more diverse profile of other data modalities, such as radiology notes, images, electrocardiograms, and even genomics once they become available. 

\hfill 

\noindent
We also note that the counterfactual outcomes are not directly observable and are estimated. A more rigorous testing of our findings should thus come from the implementation of a randomized control trial, where patients are randomly assigned to different treatment groups. For this reason, our model also cannot directly provide causality evidence. The causal relationship between valve prescription and peacemaker rate is uncertain, as it remains unclear if the valve type reflects the underlying patient characteristics, unobservable technical factor, or other unobservable confounders that can impact pacemaker implantation.

\section{Conclusion}
We provide the first study and recommendation guideline that leverages artificial intelligence-based techniques to tackle the ongoing debate regarding the optimal transcatheter valve prescription for patients with aortic stenosis. Our OPT results show the importance of conduction defect as well as other factors in prescribing the optimal valve type to reduce the risk of PPI, a clinically and operationally significant post-operative complication. These insights are derived and validated using an interpretable machine learning approach. It is important to keep in mind that the projected reductions in PPI from applying optimal valve prescriptions are statistical in nature and would benefit from real-world validation. The consistency of the results in both the train and test set in our internal dataset, as well as in a completely different external cohort, suggests the unbiased prescriptive power and the generalizability of our approach. Our study highlights the potential positive clinical benefits of deploying interpretable machine learning models into healthcare systems.

\hfill 

\noindent
\textbf{Funding Statement:}
The author(s) received no financial support for the research, authorship, and/or publication of this article.

\hfill 

\noindent
\textbf{Competing interests:} 
The author(s) has/have no competing interests to declare.

\hfill 

\noindent
\textbf{Ethics Statement:} 
All independent organizations and the Massachusetts Institute of Technology institutional review boards approved this protocol as minimal-risk research using data collected for standard clinical practice and waived the requirement for informed consent. The survey was anonymous and confidentiality of information was assured.

\hfill 

\noindent
\textbf{Data Availability:}
All datasets that are used in this study come from an academic medical center that applies to the Health Insurance Portability and Accountability Act. Due to the data protection laws, the dataset cannot be directly released to another organization. 

\newpage
\bibliography{bib}

\clearpage

\appendix

\section{Patient Characteristics} \label{app:a}

In the following section, we present summaries of patient characteristics comparing (i) the train and test set patient cohorts (see Table \ref{tab:train_test_comp}), (ii) the Medtronic Evolut and Edwards Sapien cohorts (see Table \ref{tab:valve_comp}), and (iii) the U.S. (in-sample) and Greek (out-sample) patient cohorts (see Table \ref{tab:hospital_comp}). In Table \ref{tab:missingness}, we also summarize the missingness of each patient characteristic.

\renewcommand{\thetable}{A\arabic{table}} % Prefix table numbers with 'A' in Appendix A
\setcounter{table}{0}
\renewcommand{\thefigure}{A\arabic{figure}}
\setcounter{figure}{0}
\captionsetup{
  width=0.85\linewidth,    % Set caption width to the width of the table
  skip=10pt,           % Adjust space between caption and table
  labelfont=bf,        % Make the label bold
  labelsep=period,     % Use a period after the label
  singlelinecheck=false, % Allow caption to take full width
  format=plain         % Disable the line under the caption
}
\begin{longtable}{l|c|c}
\textbf{Characteristic} & \textbf{Train, No. (\%)} & \textbf{Test, No. (\%)} \\ \hline\hline
\endfirsthead

\multicolumn{3}{c}%
{{\bfseries \tablename\ \thetable{} -- continued from previous page}} \\
\hline
\textbf{Characteristic} & \textbf{Train, No. (\%)} & \textbf{Test, No. (\%)} \\ \hline
\endhead

\multicolumn{3}{r}{{\textit{Continued on next page}}} \\
\endfoot

\endlastfoot
Sex & & \\ 
\quad Female & 435 (48.88\%) & 416 (46.79\%) \\ 
\quad Male & 455 (51.12\%) & 473 (53.21\%) \\ \hline
Weight (kg), median (IQR) & 77.1 (66.0-88.98) & 77.0 (66.0-90.9) \\ \hline
Porcelain Aorta & & \\ 
\quad False & 875 (98.31\%) & 877 (98.65\%) \\ 
\quad True & 15 (1.69\%) & 12 (1.35\%) \\ \hline
Prior Peripheral Arterial Disease & & \\ 
\quad False & 700 (78.65\%) & 706 (79.42\%) \\ 
\quad True & 190 (21.35\%) & 183 (20.58\%) \\ \hline
Transient Ischemic Attacks & & \\ 
\quad False & 842 (94.61\%) & 821 (92.35\%) \\ 
\quad True & 48 (5.39\%) & 68 (7.65\%) \\ \hline
Prior Myocardial Infarction & & \\ 
\quad False & 714 (80.22\%) & 690 (77.62\%) \\ 
\quad True & 176 (19.78\%) & 199 (22.38\%) \\ \hline
Prior CABG & & \\ 
\quad False & 764 (85.84\%) & 764 (85.94\%) \\ 
\quad True & 126 (14.16\%) & 125 (14.06\%) \\ \hline
\makecell[l]{Prior Percutaneous Coronary \\ Intervention} & & \\ 
\quad False & 658 (73.93\%) & 657 (73.9\%) \\ 
\quad True & 232 (26.07\%) & 232 (26.1\%) \\ \hline
Supplemental Home Oxygen & & \\ 
\quad False & 840 (94.38\%) & 845 (95.05\%) \\ 
\quad True & 50 (5.62\%) & 44 (4.95\%) \\ \hline
Current Dialysis & & \\ 
\quad False & 859 (96.52\%) & 872 (98.09\%) \\ 
\quad True & 31 (3.48\%) & 17 (1.91\%) \\ \hline
Hypertension & & \\ 
\quad False & 91 (10.22\%) & 83 (9.34\%) \\ 
\quad True & 799 (89.78\%) & 806 (90.66\%) \\ \hline
Left Ventricle EF, median (IQR) & 60.0 (54.0-65.0) & 60.0 (53.0-65.0) \\ \hline
Diabetes & & \\ 
\quad False & 576 (64.72\%) & 613 (68.95\%) \\ 
\quad True & 314 (35.28\%) & 276 (31.05\%) \\ \hline
Diabetes Control Therapy & & \\ 
\quad Diet & 39 (4.38\%) & 29 (3.26\%) \\ 
\quad Insulin & 108 (12.13\%) & 85 (9.56\%) \\ 
\quad No & 586 (65.84\%) & 627 (70.53\%) \\ 
\quad Oral & 157 (17.64\%) & 148 (16.65\%) \\ \hline
Age, median (IQR) & 82.0 (76.0-87.0) & 82.0 (76.0-87.0) \\ \hline
BMI, median (IQR) & 27.78 (24.26-32.03) & 27.83 (24.38-32.11) \\ \hline
Atrial Fibrillation Classification & & \\ 
\quad No & 583 (65.51\%) & 596 (67.04\%) \\ 
\quad Paroxysmal & 144 (16.18\%) & 146 (16.42\%) \\ 
\quad Persistent & 163 (18.31\%) & 147 (16.54\%) \\ \hline
\makecell[l]{Aortic Valve Area, \\ median (IQR)} & 0.7 (0.6-0.83) & 0.7 (0.59-0.81) \\ \hline
\makecell[l]{Peak Aortic Valve Gradient, \\ median (IQR)} & 69.29 (58.0-79.0) & 69.58 (58.0-81.0) \\ \hline
Carotid Artery Stenosis & & \\ 
\quad False & 707 (79.44\%) & 714 (80.31\%) \\ 
\quad True & 183 (20.56\%) & 175 (19.69\%) \\ \hline
Coronary Artery Disease & & \\ 
\quad False & 747 (83.93\%) & 758 (85.26\%) \\ 
\quad True & 143 (16.07\%) & 131 (14.74\%) \\ \hline
Chronic Lung Disease & & \\ 
\quad False & 550 (61.8\%) & 545 (61.3\%) \\ 
\quad True & 340 (38.2\%) & 344 (38.7\%) \\ \hline
Prior Aortic Valve Procedure & & \\ 
\quad False & 876 (98.43\%) & 874 (98.31\%) \\ 
\quad True & 14 (1.57\%) & 15 (1.69\%) \\ \hline
Immunosuppressive Medication & & \\ 
\quad False & 818 (91.91\%) & 797 (89.65\%) \\ 
\quad True & 72 (8.09\%) & 92 (10.35\%) \\ \hline
Prior 2 Weeks Heart Failure & & \\ 
\quad False & 272 (30.56\%) & 271 (30.48\%) \\ 
\quad True & 618 (69.44\%) & 618 (69.52\%) \\ \hline
Left Main Coronary Disease & & \\ 
\quad False & 829 (93.15\%) & 834 (93.81\%) \\ 
\quad True & 61 (6.85\%) & 55 (6.19\%) \\ \hline
\makecell[l]{Luminal Narrowing of  \\ Prox LAD $>$ 70\%} & & \\ 
\quad False & 754 (84.72\%) & 768 (86.39\%) \\ 
\quad True & 136 (15.28\%) & 121 (13.61\%) \\ \hline
Aortic Valve Annular Calcification & & \\ 
\quad False & 168 (18.88\%) & 198 (22.27\%) \\ 
\quad True & 722 (81.12\%) & 691 (77.73\%) \\ \hline
Hostile Chest & & \\ 
\quad False & 856 (96.18\%) & 846 (95.16\%) \\ 
\quad True & 34 (3.82\%) & 43 (4.84\%) \\ \hline
Conduction Defect & & \\ 
\quad False & 654 (73.48\%) & 622 (69.97\%) \\ 
\quad True & 236 (26.52\%) & 267 (30.03\%) \\ \hline

Aortic Insufficiency, No. (\%) & & \\ 
\quad None & 197 (22.13\%) & 203 (22.83\%) \\
\quad Trace/Trivial & 188 (21.12\%) & 171 (19.24\%) \\ 
\quad Mild & 393 (44.16\%) & 393 (44.21\%) \\ 
\quad Moderate & 102 (11.46\%) & 115 (12.94\%) \\ 
\quad Severe & 10 (1.12\%) & 7 (0.79\%) \\ \hline
Aortic Valve Morphology, No. (\%) & & \\ 
\quad Bicuspid & 33 (3.71\%) & 32 (3.6\%) \\ 
\quad Other & 0 (0.0\%) & 1 (0.11\%) \\
\quad Tricuspid & 778 (87.42\%) & 801 (90.1\%) \\ 
\quad Uncertain & 73 (8.2\%) & 51 (5.74\%) \\ 
\quad Unicuspid & 6 (0.67\%) & 4 (0.45\%) \\ \hline
Etiology of Aortic Valve Disease, No. (\%) & & \\ 
\quad Congenital & 9 (1.01\%) & 10 (1.12\%) \\
\quad Degenerative & 814 (91.46\%) & 820 (92.24\%) \\
\quad Other & 66 (7.42\%) & 55 (6.19\%) \\
\quad Rheumatic fever & 1 (0.11\%) & 4 (0.45\%) \\ \hline
Number of Diseased Vessels, No. (\%) & & \\ 
\quad None & 391 (43.93\%) & 427 (48.03\%) \\
\quad One & 221 (24.83\%) & 180 (20.25\%) \\
\quad Two & 121 (13.6\%) & 121 (13.61\%) \\
\quad Three & 157 (17.64\%) & 161 (18.11\%) \\ \hline
\makecell[l]{Left Ventricular Internal \\ Diastolic Dimension, median (IQR)} & 4.47 (4.3-4.5) & 4.47 (4.2-4.8) \\ \hline
\makecell[l]{Left Ventricular Internal \\ Systolic Dimension, median (IQR)} & 3.13 (2.8-3.3) & 3.13 (2.8-3.3) \\ \hline
\makecell[l]{Right Coronary Artery Height, \\ median (IQR)} & 17.61 (16.0-18.6) & 17.61 (16.1-18.6) \\ \hline
\makecell[l]{Left Coronary Artery Height, \\ median (IQR)} & 15.66 (14.7-16.3) & 15.66 (14.7-16.0) \\ \hline
Aortic Root Angle, median (IQR) & 48.58 (45.52-51.0) & 48.58 (46.0-51.0) \\ \hline
\makecell[l]{Sinotubular Junction Diameter, \\ median (IQR)} & 28.38 (27.6-29.0) & 28.38 (27.0-29.0) \\ \hline
Sinus of Valsalva Diameter, median (IQR) & 34.57 (33.4-35.7) & 34.57 (33.0-35.4) \\ \hline
\makecell[l]{Major Aortic Annulus Diameter, \\ median (IQR)} & 27.83 (27.0-28.8) & 27.83 (26.6-28.7)\\ \hline
\makecell[l]{Minor Aortic Annulus Diameter, \\ median (IQR)} & 21.9 (21.1-22.6) & 21.9 (20.6-22.6) \\ \hline
\makecell[l]{Mean Diameter of Aortic Annulus, \\ median (IQR)} & 24.73 (24.01-25.65) & 24.73 (23.75-25.55) \\ \hline
\makecell[l]{Calculated Perimeter of Aortic Annulus, \\ median (IQR)} & 77.67 (75.52-80.54) & 77.67 (74.58-80.23) \\ \hline
\makecell[l]{Measured Perimeter of Aortic Annulus, \\ median (IQR)} & 78.33 (75.6-80.8) & 78.33 (75.0-80.5) \\ \hline
Area of Aortic Annulus, median (IQR) & 4.72 (4.35-4.97) & 4.72 (4.29-4.92) \\
\hline
Valve Type, No. (\%) &  &  \\
\quad Medtronic Evolut & 224 (25.17\%) & 249 (28.01\%) \\ 
\quad Edwards Sapien & 666 (74.83\%) & 640 (71.99\%) \\ \hline
PPI Rate, median (IQR) & 0.0 (0.0-0.0) & 0.0 (0.0-0.0) \\ \hline
\caption{A summary of the train and test patient cohorts. For quantitative variables, the median and inter-quartile range of both sets is presented. For categorical variables, the number and percentage of patients assigned to each category. } \label{tab:train_test_comp} \\
\end{longtable} 

\begin{longtable}{l|c|c}

\hline
\textbf{Characteristic} & \textbf{Medtronic Evolut} & \textbf{Edwards Sapien} \\ \hline
\endfirsthead

\multicolumn{3}{c}%
{{\bfseries \tablename\ \thetable{} -- continued from previous page}} \\
\hline
\textbf{Characteristic} & \textbf{Medtronic Evolut} & \textbf{Edwards Sapien} \\ \hline
\endhead

\multicolumn{3}{r}{{\textit{Continued on next page}}} \\ 
\endfoot

\endlastfoot

Sex, No. (\%) & & \\ 
\quad Female & 275 (58.14\%) & 576 (44.1\%) \\ 
\quad Male & 198 (41.86\%) & 730 (55.9\%) \\ \hline

Weight (kg), median (IQR) & 75.0 (64.0-86.36) & 78.0 (66.5-91.0) \\ \hline

Porcelain Aorta, No. (\%) & & \\ 
\quad False & 466 (98.52\%) & 1286 (98.47\%) \\ 
\quad True & 7 (1.48\%) & 20 (1.53\%) \\ \hline

Prior Peripheral Arterial Disease, No. (\%) & & \\ 
\quad False & 378 (79.92\%) & 1028 (78.71\%) \\ 
\quad True & 95 (20.08\%) & 278 (21.29\%) \\ \hline

Transient Ischemic Attacks, No. (\%) & & \\ 
\quad False & 441 (93.23\%) & 1222 (93.57\%) \\ 
\quad True & 32 (6.77\%) & 84 (6.43\%) \\ \hline

Prior Myocardial Infarction, No. (\%) & & \\ 
\quad False & 375 (79.28\%) & 1029 (78.79\%) \\ 
\quad True & 98 (20.72\%) & 277 (21.21\%) \\ \hline

Prior CABG, No. (\%) & & \\ 
\quad False & 411 (86.89\%) & 1117 (85.53\%) \\ 
\quad True & 62 (13.11\%) & 189 (14.47\%) \\ \hline

\makecell[l]{Prior Percutaneous \\ Coronary Intervention, No. (\%)} & & \\ 
\quad False & 349 (73.78\%) & 966 (73.97\%) \\ 
\quad True & 124 (26.22\%) & 340 (26.03\%) \\ \hline

Supplemental Home Oxygen, No. (\%) & & \\ 
\quad False & 457 (96.62\%) & 1228 (94.03\%) \\ 
\quad True & 16 (3.38\%) & 78 (5.97\%) \\ \hline

Current Dialysis, No. (\%) & & \\ 
\quad False & 461 (97.46\%) & 1270 (97.24\%) \\ 
\quad True & 12 (2.54\%) & 36 (2.76\%) \\ \hline

Hypertension, No. (\%) & & \\ 
\quad False & 55 (11.63\%) & 119 (9.11\%) \\ 
\quad True & 418 (88.37\%) & 1187 (90.89\%) \\ \hline

Left Ventricle EF, median (IQR) & 60.0 (55.0-65.0) & 60.0 (53.0-65.0) \\ \hline

Diabetes, No. (\%) & & \\ 
\quad False & 317 (67.02\%) & 872 (66.77\%) \\ 
\quad True & 156 (32.98\%) & 434 (33.23\%) \\ \hline

Diabetes Control Therapy, No. (\%) & & \\ 
\quad Diet & 20 (4.23\%) & 48 (3.68\%) \\ 
\quad Insulin & 51 (10.78\%) & 142 (10.87\%) \\ 
\quad No & 324 (68.5\%) & 889 (68.07\%) \\ 
\quad Oral & 78 (16.49\%) & 227 (17.38\%) \\ \hline

Age, median (IQR) & 82.0 (77.0-88.0) & 82.0 (76.0-87.0) \\ \hline

BMI, median (IQR) & 27.6 (24.1-31.8) & 28.0 (24.4-32.2) \\ \hline

Atrial Fibrillation Classification, No. (\%) & & \\ 
\quad No & 321 (67.86\%) & 858 (65.7\%) \\ 
\quad Paroxysmal & 92 (19.45\%) & 198 (15.16\%) \\ 
\quad Persistent & 60 (12.68\%) & 250 (19.14\%) \\ \hline

Aortic Valve Area, median (IQR) & 0.71 (0.6-0.84) & 0.7 (0.59-0.81) \\ \hline

\makecell[l]{Peak Aortic Valve Gradient,\\ median (IQR)} & 68.0 (56.0-81.0) & 69.58 (58.0-80.0) \\ \hline

Carotid Artery Stenosis, No. (\%) & & \\ 
\quad False & 377 (79.7\%) & 1044 (79.94\%) \\ 
\quad True & 96 (20.3\%) & 262 (20.06\%) \\ \hline

Coronary Artery Disease, No. (\%) & & \\ 
\quad False & 396 (83.72\%) & 1109 (84.92\%) \\ 
\quad True & 77 (16.28\%) & 197 (15.08\%) \\ \hline

Chronic Lung Disease, No. (\%) & & \\ 
\quad False & 319 (67.44\%) & 776 (59.42\%) \\ 
\quad True & 154 (32.56\%) & 530 (40.58\%) \\ \hline

Prior Aortic Valve Procedure, No. (\%) & & \\ 
\quad False & 467 (98.73\%) & 1283 (98.24\%) \\ 
\quad True & 6 (1.27\%) & 23 (1.76\%) \\ \hline

Immunosupressive Medication, No. (\%) & & \\ 
\quad False & 421 (89.01\%) & 1194 (91.42\%) \\ 
\quad True & 52 (10.99\%) & 112 (8.58\%) \\ \hline

Prior 2 Weeks Heart Failure, No. (\%) & & \\ 
\quad False & 120 (25.37\%) & 423 (32.39\%) \\ 
\quad True & 353 (74.63\%) & 883 (67.61\%) \\ \hline

Left Main Coronary Disease, No. (\%) & & \\ 
\quad False & 443 (93.66\%) & 1220 (93.42\%) \\ 
\quad True & 30 (6.34\%) & 86 (6.58\%) \\ \hline

\makecell[l]{Luminal Narrowing of \\ Prox LAD $>$ 70\%, No. (\%)} & & \\ 
\quad False & 411 (86.89\%) & 1111 (85.07\%) \\ 
\quad True & 62 (13.11\%) & 195 (14.93\%) \\ \hline

\makecell[l]{Aortic Valve \\ Annular Calcification, No. (\%)} & & \\ 
\quad False & 137 (28.96\%) & 229 (17.53\%) \\ 
\quad True & 336 (71.04\%) & 1077 (82.47\%) \\ \hline

Hostile Chest, No. (\%) & & \\ 
\quad False & 459 (97.04\%) & 1243 (95.18\%) \\ 
\quad True & 14 (2.96\%) & 63 (4.82\%) \\ \hline

Conduction Defect, No. (\%) & & \\ 
\quad False & 654 (73.48\%) & 622 (69.97\%) \\ 
\quad True & 236 (26.52\%) & 267 (30.03\%) \\ \hline

Aortic Insufficiency, No. (\%) & & \\ 
\quad None & 103 (21.78\%) & 297 (22.74\%) \\
\quad Trace/Trivial & 100 (21.14\%) & 259 (19.83\%) \\ 
\quad Mild & 220 (46.51\%) & 566 (43.34\%) \\ 
\quad Moderate & 47 (9.94\%) & 170 (13.02\%) \\ 
\quad Severe & 3 (0.63\%) & 14 (1.07\%) \\ \hline

Aortic Valve Morphology, No. (\%) & & \\ 
\quad Bicuspid & 7 (1.48\%) & 58 (4.44\%) \\ 
\quad Other & 1 (0.21\%) & 0 (0.0\%) \\ 
\quad Tricuspid & 412 (87.1\%) & 1167 (89.36\%) \\ 
\quad Uncertain & 51 (10.78\%) & 73 (5.59\%) \\ 
\quad Unicuspid & 2 (0.42\%) & 8 (0.61\%) \\ \hline

Etiology of Aortic Valve Disease, No. (\%) & & \\ 
\quad Congenital & 1 (0.21\%) & 18 (1.38\%) \\ 
\quad Degenerative & 452 (95.56\%) & 1182 (90.51\%) \\ 
\quad Other & 20 (4.23\%) & 101 (7.73\%) \\ 
\quad Rheumatic fever & 0 (0.0\%) & 5 (0.38\%) \\ \hline

Number of Diseased Vessels, No. (\%) & & \\ 
\quad None & 233 (49.26\%) & 585 (44.79\%) \\ 
\quad One & 97 (20.51\%) & 304 (23.28\%) \\ 
\quad Two & 55 (11.63\%) & 187 (14.32\%) \\ 
\quad Three & 88 (18.6\%) & 230 (17.61\%) \\ \hline

\makecell[l]{Left Ventricular Internal \\ Diastolic Dimension, median (IQR)} & 4.47 (4.3-4.5) & 4.47 (4.2-4.8) \\ \hline

\makecell[l]{Left Ventricular Internal \\ Systolic Dimension, median (IQR)} & 3.13 (2.8-3.13) & 3.13 (2.8-3.5) \\ \hline

\makecell[l]{Right Coronary Artery Height, \\ median (IQR)} & 17.61 (16.0-17.61) & 17.61 (16.2-19.0) \\ \hline

\makecell[l]{Left Coronary Artery Height, \\ median (IQR)} & 15.66 (14.6-15.66) & 15.66 (14.9-16.8) \\ \hline

Aortic Root Angle, median (IQR) & 48.58 (46.0-49.0) & 48.58 (46.0-52.0) \\ \hline

\makecell[l]{Sinotubular Junction Diameter, \\ median (IQR)} & 28.38 (26.4-28.38) & 28.38 (27.6-29.25) \\ \hline

Sinus of Valsalva Diameter, median (IQR) & 34.57 (32.0-34.57) & 34.57 (33.8-36.0) \\ \hline

\makecell[l]{Major Aortic Annulus Diameter, \\ median (IQR)} & 27.83 (26.1-27.83) & 27.83 (27.0-29.2) \\ \hline

\makecell[l]{Minor Aortic Annulus Diameter, \\ median (IQR)} & 21.9 (20.2-21.9) & 21.9 (21.1-23.0) \\ \hline

\makecell[l]{Mean Diameter of Aortic Annulus, \\ median (IQR)} & 24.73 (23.2-24.73) & 24.73 (24.1-26.1) \\ \hline

\makecell[l]{Calculated Perimeter of Aortic Annulus, \\ median (IQR)} & 77.7 (72.9-77.7) & 77.7 (75.7-82.0) \\ \hline

\makecell[l]{Measured Perimeter of Aortic Annulus, \\ median (IQR)} & 78.33 (73.5-78.33) & 78.33 (76.2-82.07) \\ \hline

Area of Aortic Annulus, median (IQR) & 4.72 (4.07-4.72) & 4.72 (4.4-5.13) \\
\hline

Valve Type, No. (\%) &  &  \\
\quad Medtronic Evolut & 473 (100\%) & 0 (0.0\%) \\ 
\quad Edwards Sapien & 0 (0.0\%) & 1306 (100.00\%) \\ \hline

PPI Rate, median (IQR) & 0.0 (0.0-0.0) & 0.0 (0.0-0.0) \\ \hline

\caption{A summary of the Medtronic Evolut and Edwards Sapien patient cohorts. For quantitative variables, the median and inter-quartile range of both sets is presented. For categorical variables, the number and percentage of patients assigned to each category. }
\label{tab:valve_comp}
\end{longtable}

\clearpage
\begin{longtable}{l|c|c}
\hline
\textbf{Characteristic} & \textbf{Hospital A} & \textbf{Hospital B} \\ \hline
\endfirsthead

\multicolumn{3}{c}%
{{\bfseries \tablename\ \thetable{} -- continued from previous page}} \\
\hline
\textbf{Characteristic} & \textbf{Train, No. (\%)} & \textbf{Test, No. (\%)} \\ \hline
\endhead

\multicolumn{3}{r}{{\textit{Continued on next page}}} \\ 
\endfoot

\endlastfoot

Age, median (IQR) & 82.0 (76.0-87.0) & 81.0 (74.3-85.0) \\ \hline

Sex, No. (\%) & & \\ \hline
\quad Female & 851 (47.84\%) & 71 (40.8\%) \\ 
\quad Male & 928 (52.16\%) & 103 (59.2\%) \\ \hline

Weight (kg), median (IQR) & 77.0 (66.0-90.0) & 78.0 (68.0-89.0) \\ \hline

Conduction Defect, No. (\%) & & \\ \hline
\quad False & 1276 (71.73\%) & 147 (84.48\%) \\ 
\quad True & 503 (28.27\%) & 26 (14.94\%) \\ \hline

\makecell[l]{Minor Aortic Annulus Diameter, \\ median (IQR)} & 21.9 (20.8-22.6) & 21.0 (20.0-23.0) \\ \hline

Peak Aortic Valve Gradient, median (IQR) & 69.58 (58.0-81.0) & 70.0 (60.0-88.0) \\ \hline

\makecell[l]{Left Ventricular Internal \\ Diastolic Dimension, median (IQR)} & 4.47 (4.2-4.8) & 4.7 (4.2-5.22) \\
\hline
\caption{A comparison between the patient cohorts of U.S. Hopsital and Greek Hospital (used for external validation) with respect to the patient demographics and the variables used by the proposed OPT. For quantitative variables, the median and inter-quartile range of both sets is presented. For categorical variables, the number and percentage of patients assigned to each category.} \label{tab:hospital_comp}

\end{longtable}

\begin{longtable}{l|l}
\hline
\textbf{Characteristic} & \textbf{\% Missing} \\ \hline
\endfirsthead

\multicolumn{2}{c}%
{{\bfseries \tablename\ \thetable{} -- continued from previous page}} \\
\hline
\textbf{Characteristic} & \textbf{\% Missing} \\ \hline
\endhead

\multicolumn{2}{r}{{\textit{Continued on next page}}} \\ 
\endfoot

\endlastfoot
        Sex & 0.0 \\ 
        Weight (kg) & 0.06 \\ 
        Porcelain Aorta & 0.0 \\
        Prior Peripheral Arterial Disease & 0.0 \\
        Transient Ischemic Attacks & 0.0 \\
        Prior Myocardial Infarction & 0.17 \\
        Prior CABG & 0.0 \\
        Prior Percutaneous Coronary Intervention & 0.0 \\
        Supplemental Home Oxygen & 0.11 \\
        Current Dialysis & 0.06 \\
        Hypertension & 0.0 \\
        Left Ventricle EF & 1.57 \\
        Diabetes & 0.0 \\
        Diabetes Control Therapy & 0.0 \\
        Age & 0.0 \\
        BMI & 0.11 \\
        Atrial Fibrillation Classification & 0.67 \\
        Aortic Valve Area & 2.02 \\
        Peak Aortic Valve Gradient & 5.73 \\
        Carotid Artery Stenosis & 5.96 \\
        Coronary Artery Disease & 0.34 \\
        Chronic Lung Disease & 1.35 \\
        Prior Aortic Valve Procedure & 0.0 \\
        Immunosupressive Medication & 0.06 \\
        Prior 2 Weeks Heart Failure & 0.56 \\
        Left Main Coronary Disease & 1.01 \\
        Luminal Narrowing of Prox LAD greater than 70\% & 1.29 \\
        Aortic Valve Annular Calcifaction & 2.19 \\
        Hostile Chest & 0.06 \\
        Conduction Defect & 0.0 \\
        Aortic Insufficiency & 1.57 \\
        Aortic Valve Morphology & 1.07 \\
        Etiology of Aortic Valve Disease & 0.11 \\
        Number of Diseased Vessels & 0.51 \\
        Left Ventricular Internal Diastolic Dimension & 29.79 \\
        Left Ventricular Internal Systolic Dimension & 30.13 \\
        Right Coronary Artery Height & 32.72 \\
        Left Coronary Artery Height & 32.83 \\
        Aortic Root Angle & 32.66 \\
        Sinotubular Junction Diameter & 38.84 \\
        Sinus of Valsalva Diameter & 36.65 \\
        Major Aortic Annulus Diameter & 31.98 \\
        Minor Aortic Annulus Diameter & 32.04 \\
        Mean Diameter of Aortic Annulus & 31.59 \\
        Calculated Perimeter of Aortic Annulus & 31.82 \\
        Measured Perimeter of Aortic Annulus & 32.72 \\
        Area of Aortic Annulus & 32.77 \\ \hline
\caption{A summary of the missingness for each feature present in our dataset.} \label{tab:missingness}

\end{longtable}

\section{Additional Figures \& Results} \label{app:addit_figures}

\renewcommand{\thetable}{B\arabic{table}} % Prefix table numbers with 'B' in Appendix B
\setcounter{table}{0}
\renewcommand{\thefigure}{B\arabic{figure}} % Prefix figure numbers with 'B'
\setcounter{figure}{0} % Reset the figure counter

We include some additional figures and results not presented in the main body.

\subsection{Counterfactual Estimation} \label{app:addit_figures_counter}

\begin{figure}[hbt]
    \centering
    \includegraphics[width=0.85\linewidth]{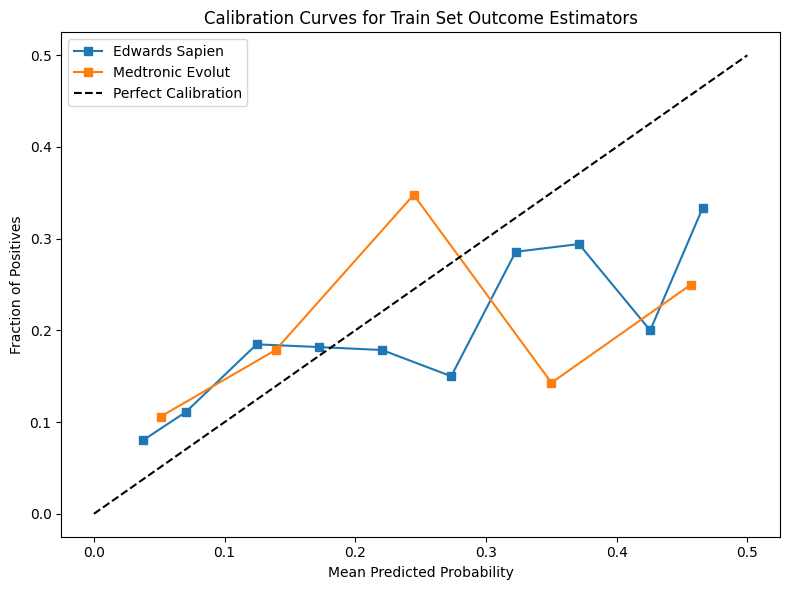}
    \caption{Calibration Curves for Training Set Outcome Estimators. We calibrate the train set outcome estimators for Edwards Sapien and Medtronic Evolut on patients in the testing cohort who received Sapien and Evolut, respectively. We leave out 2.5\% outliers on either side of the predicted probability.}
    \label{fig:train-cc}
\end{figure}

\begin{figure}[hbt]
    \centering
    \includegraphics[width=0.85\linewidth]{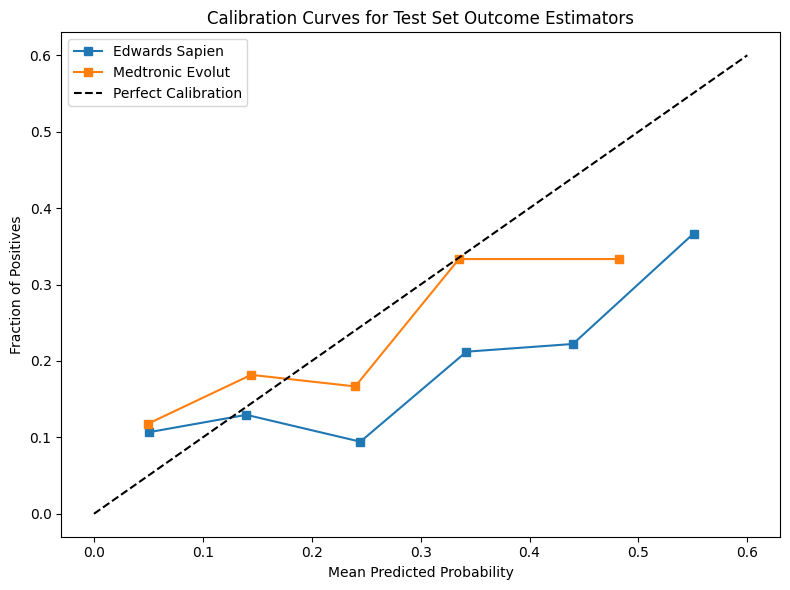}
    \caption{Calibration Curves for Testing Set Outcome Estimators. We calibrate the test set outcome estimators for Edwards Sapien and Medtronic Evolut on patients in the training cohort who received Sapien and Evolut, respectively. We leave out 2.5\% outliers on either side of the predicted probability.}
    \label{fig:test-cc}
\end{figure}

\begin{figure}[hbt]
    \centering
    \includegraphics[width=\linewidth]{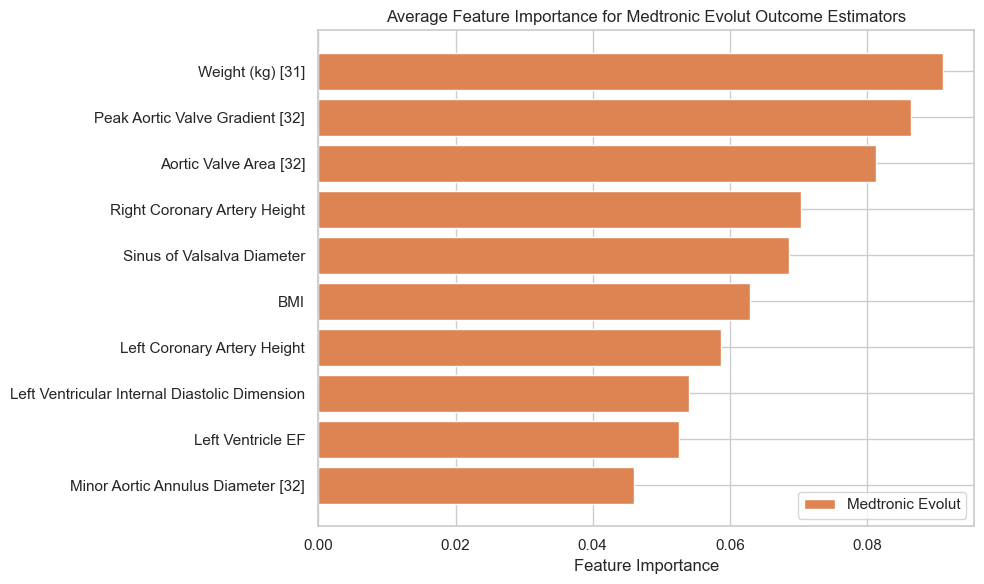}
    \caption{Evolut Feature Importance: Feature importances of the Medtronic Evolut valve outcome estimators. Importance score are averaged over the associated train and test set models.}
    \label{fig:estimate-feat-import-evolut}
\end{figure}

Table \ref{tab:reward} gives the AUC-ROC for the six XGBoost models that comprise the train and test counterfactual estimators. In Figures \ref{fig:train-cc} and \ref{fig:test-cc} we plot calibration curves for the Edwards Sapien and Medtronic Evolut outcome estimators for the training and testing sets, respectively. In particular, for Figure \ref{fig:train-cc}, we deploy the training set Edwards Sapien (Medtronic Evolut) estimator on test set patients who received a Sapien (Evolut) valve, before comparing the generated risk estimate to the actual pacemaker rate of Sapien (Evolut) patients in the test set. In Figure \ref{fig:test-cc}, we do the analogous for our testing set outcome estimators, calibrating them on the training set population who received the appropriate valve. For each calibration curve, we use 10 buckets and exclude 5\% outliers in predicted probability. Finally, Figure \ref{fig:estimate-feat-import-evolut} lists the ten most important features for the Medtronic Evolut outcome estimators.

\begin{table}[hbt]
\centering
\resizebox{0.9\textwidth}{!}{
\begin{tabular}{cccc}
\hline
  & Propensity & Outcome- Medtronic Evolut & Outcome- Edwards Sapien   \\ 
      \hline
Train Estimator & 0.6149 & 0.4674 & 0.5942  \\ 
Test Estimator &  0.6823 & 0.6288 & 0.5970 \\
\hline
\end{tabular}}
\caption{AUC Performance: The AUC-ROC of the propensity and outcome models used to estimate the counterfactual pacemaker rate. The estimations across the train and test sets are largely equivalent/close to each other. }
\label{tab:reward}
\end{table}

\subsection{Subgroup Analysis} \label{app:addit_figures_subgroup}

We divide our testing cohort into five subgroups based on patient age: 75 and younger ($N = 195$), 76 to 80 ($N = 195$), 81 to 85 ($N = 216$), 86 to 90 ($N = 193$), and 91 and older ($N = 95$). Table \ref{tab:subgroup} details the percent improvement of our model over the historical policy for each cohort using the counterfactual estimator. We elect not to use the node analysis because the proportionally smaller cohorts yield fewer patients per node, complicating the technique's extrapolation of pacemaker rates based on historical averages.

\begin{table}[hbt]
\centering
\resizebox{0.9\textwidth}{!}{
\begin{tabular}{ccc}
\hline
 Age Range & Cohort Size & Percent Improvement (CE)   \\ 
      \hline
$<75$ & 195 & 18.78 \\
$76-80$ & 195 & 19.89 \\ 
$81-85$ & 216 & 20.13  \\ 
$86-90$ & 193 & 9.77  \\ 
$91+$ & 95 & 12.55  \\ 
\hline
\end{tabular}}
\caption{Subgroup Analysis. A summary of the estimate pacemaker rate improvement of the proposed policy as evaluated by the counterfactual estimator for patients of different age ranges. We observe consistent improvement, with the model offering increased benefit to younger patients.}
\label{tab:subgroup}
\end{table}

\hfill

\noindent 
We observe consistent improvement across the five subgroups, with younger patients benefiting the most from our proposed policy. For older patients, it may be the case that post-operative complications are more a function of decreased health rather than valve choice, allowing our model less room to improve outcomes. Regardless of the potential reason, our results indicate that our model can lead to even larger improvements over historical treatment policies as TAVR is more widely adopted for use in younger patients \cite{tavr-younger-pop}.

\section{Model Training} \label{app:training}

\renewcommand{\thetable}{C\arabic{table}}
\setcounter{table}{0}
\renewcommand{\thefigure}{C\arabic{figure}}
\setcounter{figure}{0}

In this section, we offer some more information on our training pipeline, including a comparison of our imputation approach with a few others and more details on our hyperparameter selection process.

\subsection{Imputation}

To validate our imputation approach, we compare our model's sensitivity to varying imputation techniques in the test set. That is, we reconstruct our test set using (i) random forest imputation, (ii) conditional mean, conditioned on historical valve prescription, and imputation before evaluating our model's performance on each \cite{jager2021benchmark}. As seen in Table \ref{tab:impute}, our performance remains consistent on these data, especially as evaluated by our counterfactual estimator.

\begin{table}[hbt]
\centering
\resizebox{0.9\textwidth}{!}{
\begin{tabular}{ccc}
\hline
 Imputation Method & Percent Improvement & Percent Improvement (CE)   \\ 
      \hline
Mean & 21.5 & 16.92 \\
Random Forest & 21.58 & 16.57  \\ 
% K-Nearest Neighbors & 17.26 & 15.00  \\ 
Conditional Mean & 21.50 & 16.92  \\ 
% Support Vector Machine & 16.02 & 16.34  \\ 
\hline
\end{tabular}}
\caption{Imputation Method Comparison. A summary of the estimate pacemaker rate improvement of the proposed policy for different imputation strategies. For each method, we train an imputation model on the training set and apply it to the test set before evaluating our model using both the node analysis technique and counterfactual estimation (CE). Thus, we investigate the model's robustness to test sets of varying data quality.}
\label{tab:impute}
\end{table}

\subsection{Counterfactual Estimation} \label{app:alt-ce}

Our main approach for estimating counterfactuals involves training two separate outcome estimators, one on the patients that got the Edwards Sapien valve and one on the patients that got the Medtronic Evolut valve, and a propensity score estimator on the full training set of the cohort. In this section, we explore the effect of training a single outcome estimator instead of two in terms of final performance of the resulting trees. The single estimator is trained using the full training set, and the valve type is appended as an extra feature to the dataset. We report the percent improvement using node analysis, as well as counterfactual estimation, using the counterfactuals coming from both the single and the two model approaches. To ensure fairness of the comparison, we generate trees across 20 different training-test splits under the two counterfactual approaches and we compare their average performance.

\begin{table}[hbt]
\centering
\resizebox{0.99\textwidth}{!}{
\begin{tabular}{cccc}
\hline
 Counterfactual Estimation Method & Percent Improvement & Percent Improvement (CE, 1 model) & Percent Improvement (CE, 2 models)  \\ 
      \hline
Single Model & $9.04 \pm 0.37$ & $11.31 \pm 0.52$ & $7.79 \pm 0.58$\\
Two Models & $8.61 \pm 0.42$ & $11.10 \pm 0.59$ & $7.52 \pm 0.56$  \\ 
\hline
\end{tabular}}
\caption{Average performance and standard error of trees trained on 20 training test splits under different counterfactual estimation approaches, as evaluated by the node analysis and each counterfactual estimation approach. }
\label{tab:counterfactuals}
\end{table}

\hfill

\noindent
We observe that the two model approach results in comparable performance to the single model approach under every evaluation method. This indicates that employing one instead of two models does not critically affect our results and increases our confidence in the robustness of the method.

\subsection{OPT Training Hyperparameters}

Moving onto our OPT training approach, we briefly justify our choice of hyperparameters and discuss the tuning of the OPT complexity parameter. As discussed in Section \ref{subsec:opt_training}, we use a max depth of 8 and restrict each leaf to contain at least 50 samples. During the training process, we observed that trees trained with stricter requirements --- shorter depth and/or more samples in each leaf --- often result in the trivial tree of depth zero which prescribes Edwards Sapien to all patients. To avoid such issues, we fix these requirements. In addition to the max depth and minimum-leaf-size hyperparameters, our final model also utilizes a complexity parameter which is automatically selected by the OPT approach \cite{amram2022optimal}. For a detailed description of this selection process, which searches over many different hyperparameter values based on the model's performance on the validation set, we recommend Section 3 of \cite{oct-2017}.

\section{Performance of Alternative Candidate Models} \label{app:alt-models}

\renewcommand{\thetable}{D\arabic{table}}
\setcounter{table}{0}
\renewcommand{\thefigure}{D\arabic{figure}}
\setcounter{figure}{0}

\begin{figure}[hbt]
    \centering
    \includegraphics[width=0.5\linewidth]{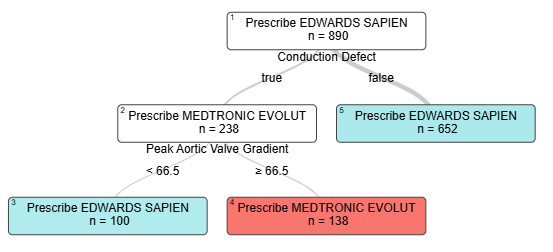}
    \caption{Alternative Candidate Model 1: An alternative high performing OPT developed during model training.}
    \label{fig:opt-0}
\end{figure}

\begin{figure}[hbt]
    \centering
    \includegraphics[width=0.5\linewidth]{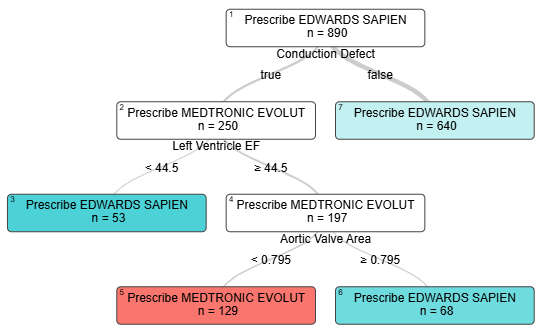}
    \caption{Alternative Candidate Model 2: An alternative high performing OPT developed during model training.}
    \label{fig:opt-24}
\end{figure}

We present the full results of our other finalists, Fin1 (Figure \ref{fig:opt-0}) and Fin2 (Figure \ref{fig:opt-24}). In particular, Tables \ref{tab:opt-0-nodes}, \ref{tab:opt-0-na}, and \ref{tab:opt-0-ce} detail Fin1's policy and performance as estimated by the node analysis technique and the counterfactual estimators. Tables \ref{tab:opt-24-nodes}, \ref{tab:opt-24-na}, and \ref{tab:opt-24-ce} do the analogous for Fin2.

% OPT-0 is Fin1
% OPT-24 is Fin2

\begin{table}[hbt]
\centering
\resizebox{\textwidth}{!}{
\begin{tabular}{c|llllll}
\multicolumn{1}{l|}{} &
    Node &
    \makecell[lc]{Prescribed Treatment} &
    \makecell[lc]{No. Implanted \\ Edwards Sapiens} &
    \makecell[lc]{Historical Sapien \\ Pacemaker Rate (\%)} &
    \makecell[lc]{No. Implanted \\ Medtronic Evoluts} &
    \makecell[lc]{Historical Evolut \\ Pacemaker Rate (\%)} \\ \hline
\multirow{3}{16mm}{Total \newline N=1779} 
    & 3 & Edwards Sapien & 155 & 18.06\% & 56 & 23.21\% \\
     & 4 & Medtronic Evolut & 224 & 31.7\% & 68 & 22.06\% \\
     & 5 & Edwards Sapien & 927 & 8.52\% & 349 & 12.61\% \\ \hline
\multirow{3}{16mm}{Train \newline N=890}
    & 3 & Edwards Sapien & 78 & 20.51\% & 22 & 27.27\% \\
     & 4 & Medtronic Evolut & 106 & 36.79\% & 32 & 25.0\% \\
     & 5 & Edwards Sapien & 485 & 8.04\% & 167 & 10.78\% \\ \hline
\multirow{3}{16mm}{Test \newline N=889} 
    & 3 & Edwards Sapien & 77 & 15.58\% & 34 & 20.59\% \\
     & 4 & Medtronic Evolut & 118 & 27.12\% & 36 & 19.44\% \\
     & 5 & Edwards Sapien & 442 & 9.05\% & 182 & 14.29\% \\ \hline
\end{tabular}
}
\caption{Historical Outcome Performance: The size and PPI rate of the historical Edwards Sapien and Medtronic Evolut patient implants in (i) the combined train and test set, (ii) the train set, and (iii) the test set for each node, as well as the treatment prescribed by Alternative Candidate Model 1 (see Figure \ref{fig:opt-0}).}
\label{tab:opt-0-nodes}
\end{table}

\begin{table}[hbt]
\centering
\resizebox{0.9\textwidth}{!}{
\begin{tabular}{cccc}
\hline
       &  Actual Pacemaker Rate  &  Observed Pacemaker Rate under OPT  &  Percent Improvement  \\
\hline
 Total &         14.05\%          &               11.88\%                &        15.49         \\
 Train &         14.16\%          &               12.07\%                &        14.73        \\
 Test  &         13.95\%          &               11.67\%                &        16.36         \\
\hline
\end{tabular}
}
\caption{Pacemaker Rate Improvement: The historical pacemaker rate of the total patient population, the train set, and the test set compared to the pacemaker rate observed under the policy prescribed by Alternative Candidate Model 1 (see Figure \ref{fig:opt-0}).}
\label{tab:opt-0-na}
\end{table}

\begin{table}[hbt]
\centering
\resizebox{0.9\textwidth}{!}{
\begin{tabular}{cccc}
\hline
       &  Actual Pacemaker Rate  &  Observed Pacemaker Rate under OPT  &  Percent Improvement  \\
\hline
 Train &         14.16\%          &               11.71\%                &        17.28         \\
 Test  &         13.95\%          &               11.17\%                &         19.9         \\
\hline
\end{tabular}
}
\caption{Pacemaker Rate Improvement (CE): The historical pacemaker rate of the training and test sets compared to the pacemaker rate observed under the policy prescribed by Alternative Candidate Model 1 (see Figure \ref{fig:opt-0}) as evaluated by the counterfactual estimators.}
\label{tab:opt-0-ce}
\end{table}

\begin{table}[hbt]
\centering
\resizebox{\textwidth}{!}{
\begin{tabular}{c|llllll}
\multicolumn{1}{l|}{} &
    Node &
    \makecell[lc]{Prescribed Treatment} &
    \makecell[lc]{No. Implanted \\ Edwards Sapiens} &
    \makecell[lc]{Historical Sapien \\ Pacemaker Rate (\%)} &
    \makecell[lc]{No. Implanted \\ Medtronic Evoluts} &
    \makecell[lc]{Historical Evolut \\ Pacemaker Rate (\%)} \\ \hline
\multirow{4}{16mm}{Total \newline N=1779} 
    & 3 & Edwards Sapien & 86 & 11.63\% & 21 & 23.81\% \\
    & 5 & Medtronic Evolut & 189 & 33.33\% & 66 & 13.64\% \\
    & 6 & Edwards Sapien & 104 & 25.0\% & 37 & 37.84\% \\
    & 7 & Edwards Sapien & 927 & 8.52\% & 349 & 12.61\% \\ \hline     
\multirow{4}{16mm}{Train \newline N=890}
    & 3 & Edwards Sapien & 43 & 16.28\% & 10 & 30.0\% \\
    & 5 & Medtronic Evolut & 95 & 40.0\% & 34 & 11.76\% \\
    & 6 & Edwards Sapien & 48 & 18.75\% & 20 & 35.0\% \\
    & 7 & Edwards Sapien & 470 & 7.66\% & 170 & 11.18\% \\ \hline
\multirow{4}{16mm}{Test \newline N=889} 
    & 3 & Edwards Sapien & 43 & 6.98\% & 11 & 18.18\% \\
    & 5 & Medtronic Evolut & 94 & 26.6\% & 32 & 15.62\% \\
    & 6 & Edwards Sapien & 56 & 30.36\% & 17 & 41.18\% \\
    & 7 & Edwards Sapien & 457 & 9.41\% & 179 & 13.97\% \\ \hline
\end{tabular}
}
\caption{Historical Outcome Performance: The size and PPI rate of the historical Edwards Sapien and Medtronic Evolut patient implants in (i) the combined train and test set, (ii) the train set, and (iii) the test set for each node, as well as the treatment prescribed by Alternative Candidate Model 2 (see Figure \ref{fig:opt-24}).}
\label{tab:opt-24-nodes}
\end{table}

\begin{table}[hbt]
\centering
\resizebox{0.9\textwidth}{!}{
\begin{tabular}{cccc}
\hline
       &  Actual Pacemaker Rate  &  Observed Pacemaker Rate under OPT  &  Percent Improvement  \\
\hline
 Total &         14.05\%          &               10.75\%                &        23.52       \\
 Train &         13.82\%          &                9.62\%                &        30.43       \\
 Test  &         14.29\%          &               11.86\%                &        16.96       \\
\hline
\end{tabular}
}
\caption{Pacemaker Rate Improvement: The historical pacemaker rate of the total patient population, the train set, and the test set compared to the pacemaker rate observed under the policy prescribed by Alternative Candidate Model 2 (see Figure \ref{fig:opt-24}).}
\label{tab:opt-24-na}
\end{table}

\begin{table}[hbt]
\centering
\resizebox{0.9\textwidth}{!}{
\begin{tabular}{cccc}
\hline
       &  Actual Pacemaker Rate  &  Observed Pacemaker Rate under OPT  &  Percent Improvement  \\
\hline
 Train &         13.82\%          &               10.08\%                &        27.06         \\
 Test  &         14.29\%          &               12.43\%                &        13.02         \\
\hline
\end{tabular}
}
\caption{Pacemaker Rate Improvement (CE): The historical pacemaker rate of the training and test sets compared to the pacemaker rate observed under the policy prescribed by Alternative Candidate Model 2 (see Figure \ref{fig:opt-24}) as evaluated by the counterfactual estimators.}
\label{tab:opt-24-ce}
\end{table}

\end{document}